\documentclass[10pt,twocolumn,letterpaper]{article}

\usepackage{cvpr}              %

\usepackage{graphicx}
\usepackage{amsmath}
\usepackage{amssymb}
\usepackage{booktabs}
\usepackage{amsfonts}       %
\usepackage{nicefrac}       %
\usepackage{microtype}      %
\usepackage[dvipsnames]{xcolor}
\usepackage{times}
\usepackage{epsfig}
\usepackage{subcaption}
\usepackage[export]{adjustbox}
\usepackage{makecell}
\usepackage{enumitem}
\usepackage{bm}
\usepackage{wrapfig}
\usepackage{xfrac}
\usepackage{sidecap}
\usepackage{float}
\usepackage{makecell}
\usepackage{stfloats}

\usepackage[pagebackref,breaklinks,colorlinks]{hyperref}

\def\vcombine{{aggregate}}
\def\vcombines{{aggregates}}
\def\vcombining{{aggregating}}
\def\vcombination{{aggregation}}
\def\heav{{HAAV}}

\usepackage[capitalize]{cleveref}
\crefname{section}{Sec.}{Secs.}
\Crefname{section}{Section}{Sections}
\Crefname{table}{Table}{Tables}
\crefname{table}{Tab.}{Tabs.}

\def\cvprPaperID{4372} %
\def\confName{CVPR}
\def\confYear{2023}

\definecolor{gold}{RGB}{218,165,32}

\begin{document}

\title{\heav{}: Hierarchical Aggregation of Augmented Views for Image Captioning}

\author{
Chia-Wen Kuo \\
Georgia Tech \\
\texttt{albert.cwkuo@gatech.edu} \\
\and
Zsolt Kira \\
Georgia Tech \\
\texttt{zkira@gatech.edu}
}

\maketitle

\begin{abstract}
A great deal of progress has been made in image captioning, driven by research into how to encode the image using pre-trained models.
This includes visual encodings (e.g. image grid features or detected objects) and more recently textual encodings (e.g. image tags or text descriptions of image regions).
As more advanced encodings are available and incorporated, it is natural to ask: how to \textbf{efficiently} and \textbf{effectively} leverage the heterogeneous set of encodings?
In this paper, we propose to regard the encodings as augmented views of the input image.
The image captioning model encodes each view independently with a shared encoder efficiently, and a contrastive loss is incorporated across the encoded views in a novel way to improve their representation quality and the model's data efficiency.
Our proposed hierarchical decoder then adaptively weighs the encoded views according to their effectiveness for caption generation by first \vcombining{} within each view at the token level, and then across views at the view level.
We demonstrate significant performance improvements of +5.6\% CIDEr on MS-COCO and +12.9\% CIDEr on Flickr30k compared to state of the arts,
and conduct rigorous analyses to demonstrate the importance of each part of our design. 
\end{abstract}
\section{Introduction}\label{sec:intro}

\begin{figure}
\centering
\includegraphics[width=\linewidth]{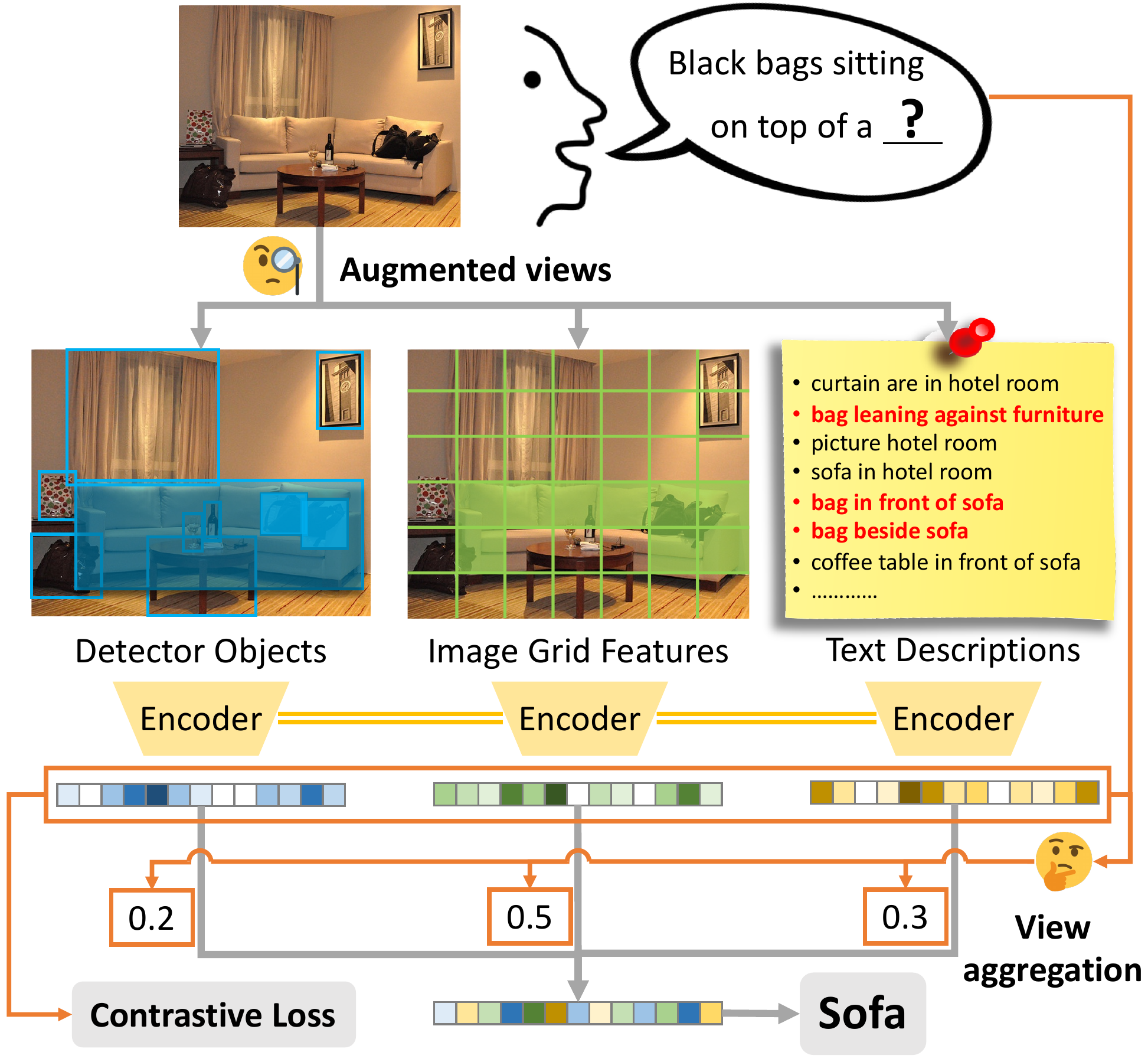}
\caption{\heav{}, \textbf{H}ierarchical \textbf{A}ggregation of \textbf{A}ugmented \textbf{V}iews, for image captioning at the step of predicting the word ``\textit{sofa}''.
First, heterogeneous views such as detected objects~\cite{anderson2018bottom}, image grid features~\cite{shen2021much}, and text descriptions~\cite{kuo2022pretrained} are generated from the input image by existing methods.
We propose to regard these views as augmentations of the input image, and independently encode each view by a shared transformer encoder efficiently.
A contrastive loss is incorporated to improve the representation quality of heterogeneous views.
Finally, our proposed hierarchical decoder models the effectiveness of each view and adaptively weigh them according to their effectiveness for predicting the current word ``\textit{sofa}''.\looseness=-1
}\label{fig:concept}
\end{figure}

A large amount of progress has been made in vision-and-language (VL) tasks
such as image captioning~\cite{chen2015microsoft,agrawal2019nocaps}, visual question answering~\cite{goyal2017making,hudson2019gqa}, and image-text retrieval.
For these tasks, recent methods~\cite{zhang2021vinvl,li2020oscar,shen2021much,kuo2022pretrained} observe that encoding the input image by an object detector~\cite{ren2015faster} pre-trained on Visual Genome~\cite{krishna2017visual} into a set of detected objects is not sufficient.
To provide information complementary to detected objects, recent works proposed to encode an input image by different pre-trained models and into different modalities, and achieve substantial performance improvement by combining these heterogeneous encodings.
For example, some works encode from the visual perspective (\textit{e.g.} stronger object detector pre-trained on a larger vocabulary and datasets~\cite{zhang2021vinvl} or global image features~\cite{ji2021improving}), while other works encode from the textual perspective (\textit{e.g.} image tags~\cite{li2020oscar} and text descriptions of image regions~\cite{kuo2022pretrained}).

Given the great success of incorporating various heterogeneous encodings or ``\textit{views}'', one research question emerges naturally: \textit{how to \textbf{efficiently} and \textbf{effectively} leverage these heterogeneous views for caption generation?}
For \textbf{efficiency}, three factors are particularly important: computation, parameter count, and label efficiency.
State-of-art VL and image captioning models are typically a transformer encoder-decoder model~\cite{vaswani2017attention}, which has undesirable quadratic computational complexity with respect to the input sequence size.
Therefore, as more views are incorporated, each represented by a sequence of tokens, we should carefully manage the computation and model size.
Moreover, on the medium-scale MS-COCO image captioning benchmark~\cite{chen2015microsoft} ($\sim$0.6M training samples), we should take label efficiency into consideration when training the data-hungry~\cite{dosovitskiy2020vit} transformer model to avoid negative effects such as overfitting.
For \textbf{effectiveness}, different views contain some shared and some complementary information of the input image.
Therefore, it is important to model the effectiveness of views and adaptively weigh them according to their effectiveness for predicting each word.
Take image captioning in Figure~\ref{fig:concept} as an example, when predicting the current word ``\textit{sofa}'', for the incomplete caption ``\textit{black bags sitting on top of a} \underline{ ? }'', if say, the view of detected objects fails to detect sofa in the input image, the captioning model should down-weigh the less effective view of detected objects and rely on other more effective views that properly encode the information about sofa.
With these considerations in mind, we propose \heav{}, \textbf{H}ierarchical \textbf{A}ggregation of \textbf{A}ugmented \textbf{V}iews.
In \heav{}, given a set of heterogeneous views of the input image from existing works such as detected objects~\cite{anderson2018bottom}, image grid features~\cite{shen2021much}, and text descriptions~\cite{kuo2022pretrained}, we propose to (1) regard heterogeneous views as augmentations of the input image, and (2) devise a hierarchical decoder layer to account for the effectiveness of heterogeneous views.

For (1), by regarding views as augmentations, we naturally choose to use a \textit{shared} transformer encoder to encode each view \textit{independently}.
Compared to methods that concatenate all views into a long sequence as input~\cite{li2020oscar,zhang2021vinvl,kuo2022pretrained}, where the computational complexity scales up quadratically with respect to the number of views, our method scales up linearly.
Compared to models that use entire models per view~\cite{huang2019attention,cornia2020meshed,rennie2017self,jiang2018recurrent} or methods that encode each view with unshared encoders~\cite{li2021align,akbari2021vatt,hu2021unit,alayrac2020self}, our method is more parameter efficient.
Furthermore, data augmentation increases data diversity and thus improves data efficiency, which is particularly important for training data-hungry transformer models.
Last but not least, by regarding heterogeneous views generated by different pre-trained models as augmentations of the input image, we incorporate a contrastive loss in a novel way to help representation learning of heterogeneous views and increase data efficiency~\cite{grill2020bootstrap,chen2020big,goyal2021self}.
Different from how other VL methods~\cite{Radford2021LearningTV,jia2021scaling,yu2022coca,li2021align,yang2022vision} incorporate a contrastive loss, our formulation does \textbf{not} require annotated pairs (\textit{e.g.} human annotated image-caption pairs in MS-COCO or image-text pairs scraped from the internet~\cite{sharma2018conceptual}) and can work with unlabeled image-only data to achieve better performance.

Also crucially, for (2) we devise a hierarchical decoder layer, which modifies the standard transformer decoder layer by introducing two-tiered cross-attention modules.
The hierarchical decoder first \vcombines{} \textit{within} each view at the token level and then \vcombines{} \textit{across} views at the view level.
By introducing this hierarchical \vcombining{} structure, we can better model the effectiveness of views and adaptively weigh them according to their effectiveness.
For example, in Section~\ref{sec:exp} Experiment, we show that if we add noise to a certain view or mask out a prominent region of the input image, the proposed hierarchical decoder indeed down-weigh the noised and masked view when generating words and captions.

To sum up, in this paper, given a set of heterogeneous views of the input image from existing works, we focus on how to \textbf{efficiently} and \textbf{effectively} leverage these views and make the following contributions:
(1) regard heterogeneous views as augmentations of the input image and propose a novel use of contrastive loss to improve computation, parameter, and data efficiency;
(2) devise a hierarchical decoder layer to model the effectiveness of each view and weigh each view accordingly for caption generation;
(3) achieve significant improvement of +5.6\% CIDEr on MS-COCO over state of the art,
and achieve comparable or often better performance compared with methods using large-scale transformer pre-training even though we do not do so;
and (4) provide thorough ablations and rigorous analyses to validate our proposed method for efficiency and effectiveness.
\section{Related Works}\label{sec:related_works}

\textbf{Image captioning.}
The goal of image captioning is to generate text descriptions for an image.
It can be roughly divided into two settings: (1) trained-from-scratch and (2) with pre-training, depending on whether the model is pre-trained on a large image-text corpus or not.
For the trained-from-scratch setting~\cite{anderson2018bottom,huang2019attention,cornia2020meshed,rennie2017self,jiang2018recurrent,pan2020x,dou2022an,li2022blip,kim2021vilt,yang2022vision}, researchers train the model only on the image captioning dataset such as MS-COCO~\cite{lin2014microsoft,chen2015microsoft}.
They mainly focus on model architecture improvements and/or better training losses, and tackle the image captioning task specifically.
On the other hand, for the pre-training setting~\cite{li2020oscar,zhang2021vinvl,chen2020uniter,tan2019lxmert,lu2019vilbert,su2019vl,li2019visualbert,li2021align,hu2021scaling,wang2022unifying,zeng2021multi}, researchers first pre-train the model on a large image-text corpus and then fine-tune to various downstream VL tasks.
They mainly focus on how to effectively pre-train the model so that it transfers well to a broad set of downstream VL tasks.
In this paper, we work on the trained-from-scratch setting for image captioning specifically.

\textbf{Image encodings.} 
One important aspect of VL approaches is to properly encode relevant information from the input image, on which the captioning model is conditioned for captions generation.
\cite{anderson2018bottom} proposed to encode finer-grained information of the input image into a sequence of objects detected by an object detector pre-trained on Visual Genome~\cite{krishna2017visual}.
This method achieved great success and soon became the dominant approach in many VL tasks~\cite{li2020oscar,chen2020uniter,tan2019lxmert,lu2019vilbert,su2019vl,li2019visualbert}.
However, recent works found that the object detector pre-trained on visual genome fails to encode important information for image captioning.
To tackle this problem VinVL~\cite{zhang2021vinvl} proposed to train the object detector on much larger training corpora that combine multiple publicly annotated object detection datasets.
Thanks to advances in foundation models~\cite{kolesnikov2020big,jia2021scaling,Radford2021LearningTV} trained on large-scale datasets, some works~\cite{li2021align,shen2021much,wang2022simvlm,dou2022an} propose to encoded the input image by a foundation model into image grid features and achieved good performance.

Nevertheless, these works still encode an input image from a single point of view by a single pre-trained model, and may still fail to encode some important information for image captioning.
For example, a pre-trained object detector may fail to encode important information such as scene information or object relationships~\cite{kuo2022pretrained}.
To encode complementary information, recent works proposed to include other heterogeneous ``\textit{views}'' of the input image along with the detected objects.
For example, OSCAR~\cite{li2020oscar} use image-level object tags along with detected object for VL pre-training.
Some works~\cite{su2019vl,Yu2021ERNIEViLKE,zhang2021rstnet,ji2021improving} include an extra token of global image features encoded by an ImageNet pre-trained image encoder.
Xmodal-Ctx~\cite{kuo2022pretrained} include text descriptions of the input image and image sub-regions by CLIP cross-modal retrieval.
In this paper, we use heterogeneous views from existing works including detected objects, image grid features, and text descriptions, and focus on how to leverage these views efficiently and effectively.

\textbf{Heterogeneous view image captioning.}
To leverage heterogeneous views encoded by different pre-trained models and potentially encoded into different modalities, common approaches include (1) a separate model for each view~\cite{huang2019attention,cornia2020meshed,rennie2017self,jiang2018recurrent}, which trains $|V|$ (number of views) models and average the predicted word probability from each model, and (2) concatenation of views~\cite{li2020oscar,zhang2021vinvl,kuo2022pretrained}, which concatenates the views, each represented by a sequence of tokens, into a long single view, and feeds the concatenate view into an image captioning model for caption generation.
However, having a separate model per view is parameter inefficient as the number of models and parameters grows linearly with $|V|$.
Concatenated views are computationally inefficient as the computational complexity from the transformer model is quadratic with respect to $|V|$.
In this paper, in pursuit of efficiency and effectiveness, we encode the views independently with a shared encoder, and devise a novel hierarchical decoder to first \vcombine{} within each view at the token level and then \vcombine{} across views at the view level.\looseness=-1

\textbf{Representation learning.}
To properly encode heterogeneous views, existing methods~\cite{li2021align,akbari2021vatt,hu2021unit,alayrac2020self} typically encode each view with a dedicated encoder.
This is parameter inefficient as the number of encoders grows linearly with respect to $|V|$.
This may also not be as data efficient and be prone to overfitting on relatively smaller-scale datasets such as  MS-COCO.
To learn a better representation of heterogeneous views, we propose to incorporate a contrastive loss, which facilitates superior self-/un-supervised representation learning~\cite{chen2020mocov2,grill2020bootstrap,tian2020makes,chen2020simple}, and is beneficial in low-label settings~\cite{grill2020bootstrap,chen2020big,goyal2021self}.
Driven by the contrastive loss, the representation of views (augmentations) from the same image are pulled together in an embedding space while that from different images are pushed apart.
Different from existing multi-modal works~\cite{Radford2021LearningTV,jia2021scaling,yu2022coca,li2021align,yang2022vision} that incorporate contrastive loss only in the pre-training stage, or require annotated pairs (\textit{e.g.} human annotated image-caption pairs in MS-COCO), our method incorporates contrastive loss together with the target image captioning task, and requires \textbf{no} annotated pairs.\looseness=-1

\section{Method}\label{sec:method}

\begin{figure*}
\centering
\includegraphics[width=0.85\linewidth]{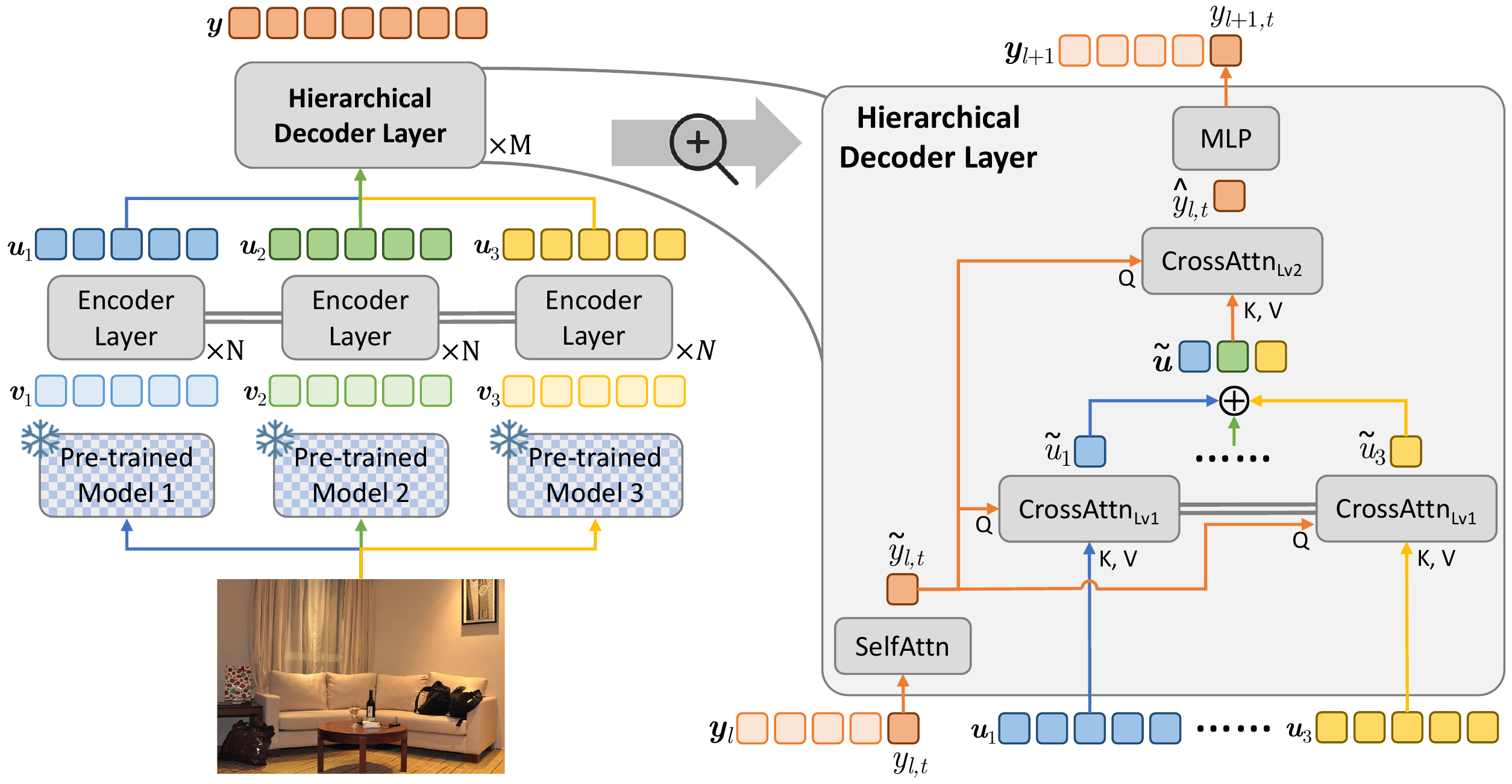}
\caption{
Model architecture.
\textbf{\textit{(Left)}} An input image is first encoded by some frozen pre-trained models into a set of heterogeneous views $\{\bm{v}_1, \bm{v}_2, \bm{v}_3\}$, each represented by a sequence of $d$-dimensional tokens $\square$.
These input views $\bm{v}$ are regarded as augmentations of the input image, and thus are encoded independently with a shared transformer encoder into $\{\bm{u}_1, \bm{u}_2, \bm{u}_3\}$.
\textbf{\textit{(Right)}} To model the effectiveness of the encoded views $\bm{u}$ for current word prediction, we propose a hierarchical decoder layer that first \vcombines{} within each encoded view $\bm{u}$ at the token level with a shared $\text{CrossAttn}_\text{Lv1}$ module, and then \vcombines{} across the encoded views at the view level with a $\text{CrossAttn}_\text{Lv2}$ module.
For clarity of illustration, we only show operations for the current $y_{l,t}$ token and not the $y_{l,1:t-1}$ past tokens.
}
\label{fig:model}
\end{figure*}

\subsection{Overview}

In image captioning, given heterogeneous views of the input image from existing works (see Figure~\ref{fig:concept}) such as detected objects~\cite{anderson2018bottom}, image grid features~\cite{shen2021much}, and text descriptions~\cite{kuo2022pretrained}, we aim to \textbf{efficiently} and \textbf{effectively} leverage these views for the target image captioning (IC) task.
Following the successful paradigm of image captioning~\cite{anderson2018bottom,cornia2020meshed,kuo2022pretrained}, an input image $\bm{x}$ is first encoded by frozen pre-trained models into heterogeneous views $\{\bm{v}_1, \bm{v}_2,...,\bm{v}_{|V|}\}$, each $\bm{v}$ represented by a sequence of $d$-dimensional tokens.
These views $\bm{v}$ are then fed into an IC transformer encoder-decoder model, where the IC-encoder encodes important information from $\bm{v}$ into $\bm{u}$ for the later decoding step, and the IC-decoder generates captions $\bm{y}$ conditioned on the encoded views $\bm{u}$.
In the rest of the paper, since the input image is encoded into $\bm{v}$ by some frozen pre-trained models from existing works and is not the focus of this paper, for the sake of brevity, we use ``\textit{input views}'' or simply ``\textit{views}'' for $\bm{v}$, ``\textit{encoded views}'' for $\bm{u}$, and ``\textit{encode}'' for the encoding process by the IC-encoder.

In Section~\ref{sec:heterogeneous-view} we revisit other successful paradigms in heterogeneous view image captioning: methods that use a separate entire model per view~\cite{huang2019attention,cornia2020meshed,rennie2017self,jiang2018recurrent} and concatenation of views~\cite{li2020oscar,zhang2021vinvl,kuo2022pretrained}.
By comparing the pros and cons of these two methods, we aim to propose our method that can efficiently account for the effectiveness of input heterogeneous views for caption generation.
In Section~\ref{sec:encoding}, we first propose to regard input views $\bm{v}$ as augmentations of the input image, and thus naturally choose to encode each $\bm{v}$ \textit{independently} with a \textit{shared} IC-encoder into $\bm{u}$.
This formulation is more efficient in terms of computation, parameter, and label.
It also allows us to add a contrastive loss in a novel way to help representation learning of the encoded views $\bm{u}$, and increase label efficiency.
In Section~\ref{sec:decoding}, we then propose a \textit{hierarchical decoder layer} to first \vcombine{} \textit{within} each encoded view $\bm{u}$ at the token level and then \vcombine{} \textit{across} all $\bm{u}$ at the view level.
By introducing this hierarchical \vcombining{} structure, the decoder better models the effectiveness of encoded views $\bm{u}$, and adaptively weighs them according to their effectiveness.
The overall model architecture is illustrated in Figure~\ref{fig:model}.

\subsection{Heterogeneous Views Image Captioning} \label{sec:heterogeneous-view}

We start by reviewing two closely related formulations in existing works: (1) combining separate models for each view~\cite{huang2019attention,cornia2020meshed,rennie2017self,jiang2018recurrent} and (2) concatenation of input views~\cite{li2020oscar,zhang2021vinvl,kuo2022pretrained}.
For (1), $|V|$ captioning models are trained independently for each input view.
The word probability predicted by these models is averaged as the final prediction.
For (2), all input views, each represented by a sequence of tokens, are concatenated along the sequence dimension to form a single long view.
The concatenated view is then fed into a single image captioning model for caption generation.

These two formulations are two extreme cases.
In (1), views are encoded and decoded independently, and the effectiveness of each view is not accounted for to generate the final prediction.
The computational complexity scales linearly, but the number of parameters also scales linearly with respect to $|V|$.
In contrast, in (2), views are encoded and decoded jointly, and the effectiveness of views for caption generation is modeled internally by the attention module within the encoder and decoder layers.
The number of parameters remains constant, but this formulation has undesirable quadratic computational complexity with respect to $|V|$.
We ask: \textit{can we find a middle ground between these two formulations that enjoys the advantages of both?}
Specifically, we would like to model the effectiveness of views for caption generation in (2), but at the same time maintain linear computational complexity in (1) and a constant number of parameters in (2).\looseness=-1

\subsection{Heterogeneous View Encoding}\label{sec:encoding}

We first observe that each input view $\bm{v}$ is generated from the input image $\bm{x}$ and thus contains partial information of $\bm{x}$.
This means that input views $\bm{v}$ can be regarded as augmentations of the input image $\bm{x}$~\cite{tian2020makes} that describe $\bm{x}$ from different perspectives.
Just like how we use data augmentation in computer vision, we would not concatenate all of the augmented images into one single image before sending it into the model or use different encoders to encode different augmented images.
Therefore, we propose to decouple the roles of the IC-encoder and the IC-decoder.
Each $\bm{v}$ is encoded \textit{independently} using a \textit{shared} IC-encoder into $\bm{u}$ efficiently for the later decoding step.
The IC-decoder then models the effectiveness of encoded views $\bm{u}$ and adaptively weighs them according to their effectiveness.
We will focus on the IC-encoder in this section and the IC-decoder in the next section.\looseness=-1 

By encoding each view independently, the computational complexity scales up linearly ($\mathcal{O}(|V|)$), which is computationally efficient.
By encoding each view with a shared IC-encoder, the model size stays constant ($\mathcal{O}(1)$), which is parameter efficient.
Data augmentation also increases data diversity and thus improves label efficiency.
Furthermore, since we regard heterogeneous input views as augmentations of the input image, we propose to add a contrastive loss~\cite{he2019moco,chen2020mocov2,chen2020simple,grill2020bootstrap} to improve the representation quality of encoded heterogeneous views $\bm{u}$.
In contrastive learning, augmented views from the same input image are positive pairs and those from different images are negative pairs.
The representation of positive pairs is pulled together in an embedding space while the representation of negative pairs is pushed apart.
To further increase diversity across heterogeneous views for the contrastive loss~\cite{chen2020simple,tian2020makes,sohn2020fixmatch,kuo2020featmatch}, we add channel-wise and sequence-wise dropout during training.
For channel-wise dropout, each channel across views and sequences is randomly zero-ed out with probability $p_c$.
For sequence-wise dropout, each token, which is a $d$-dimensional feature vector, across views is randomly zero-ed out with probability $p_s$.
Note that our goal is not to propose a new contrastive loss here but to incorporate an existing one in a novel way in the VL setting.
Compared to how most other VL pre-training methods incorporate a contrastive loss~\cite{Radford2021LearningTV,jia2021scaling,yu2022coca,li2021align,yang2022vision}, where paired annotations such as image and text pairs annotated by human or scraped from the internet are required, our formulation only requires input image \textit{without} paired annotations to construct positive and negative pairs.\looseness=-1

\subsection{Heterogeneous View Decoding}\label{sec:decoding}

Given the encoded views $\bm{u}$ from the IC-encoder, in this section we shift our attention to the IC-decoder to account for the effectiveness of $\bm{u}$ for predicting each word.
Specifically, the IC-decoder should adaptively weigh each encoded view $\bm{u}$ according to their effectiveness at each word prediction step conditioned on previously generated words and all other encoded views $\bm{u}$.
For example, in Figure~\ref{fig:concept}, when predicting the current word ``\textit{sofa}'' for the incomplete sentence ``\textit{Black bags sitting on top of a} \underline{ ? }'', if say, the object detector fails to detect sofa, the decoder should down-weigh the view of detected objects and should leverage other views that properly encode sofa information to correctly predict the next word.
On the other hand, we spent great efforts in Section~\ref{sec:encoding} to encode the views efficiently.
Therefore, the proposed decoding strategy should be at least as efficient as the proposed encoding strategy.

We propose a \textbf{hierarchical decoder layer} shown in Figure~\ref{fig:model}, which modifies the standard transformer decoder layer by introducing the two-tiered cross-attention structure.
First, current word $y_{l,t}$ is contextualized with all previously generated words $y_{l,1:t-1}$ into $\tilde{y}_{l,t}$ by the SelfAttn module.
Next, to model the effectiveness of each encoded view for current word prediction, the hierarchical decoder \vcombines{} \textit{within} each encoded view $\bm{u}$ independently at the token level by a shared $\text{CrossAttn}_\text{Lv1}$ module.
The outputs of $\text{CrossAttn}_\text{Lv1}$ from all views are collected into $\tilde{\bm{u}}$.
Finally, to adaptively weigh all the views according to their effectiveness for current word prediction, the hierarchical decoder then \vcombines{} \textit{across} views $\tilde{\bm{u}}$ at the view level by the $\text{CrossAttn}_\text{Lv2}$ module.
We also add a view-wise dropout to randomly drop an encoded view $\bm{u}$ with probability $p_v$ during training to encourage the decoder to better leverage all encoded views instead of focusing on a few certain ones
By introducing this hierarchical \vcombination{} structure, we can better model the effectiveness of the encoded views $\bm{u}$ by $\text{CrossAttn}_\text{Lv1}$ and adaptively weigh them by $\text{CrossAttn}_\text{Lv2}$.

In terms of computation and parameter efficiency, the hierarchical decoder layer uses a shared $\text{CrossAttn}_\text{Lv1}$ module to first \vcombine{} within each view.
The computational complexity scales up linearly, and the model size is constant with respect to $|V|$.
It then \vcombines{} across views using a $\text{CrossAttn}_\text{Lv2}$ module with linear computational complexity, and constant model size with respect to $|V|$.
Overall, the computational complexity and model size does not exceed our proposed view encoding strategy in Section~\ref{sec:encoding}.

\section{Experiment} \label{sec:exp}

\subsection{Implementation Details}

Our proposed \heav{} can be easily incorporated into existing transformer encoder-decoder IC models.
In this paper, we choose the state-of-art Xmodal-Ctx~\cite{kuo2022pretrained} as our base model.
Following the conventional training procedure~\cite{cornia2020meshed,li2020oscar,vinyals2015show}, the model is first trained with cross-entropy loss and then fine-tuned with SCST~\cite{rennie2017self} loss, which optimizes CIDEr score by reinforcement learning (note this is done by all compared methods).
The dropout rates $p_c$, $p_s$, and $p_t$ are all set to 0.1 following \cite{gao-etal-2021-simcse} without tuning.
We reuse the heterogeneous views from previous methods without modifications, including detected objects from \cite{anderson2018bottom}, CLIP ViT-B/32~\cite{Radford2021LearningTV} image grid features from \cite{shen2021much}, and text descriptions of image regions from \cite{kuo2022pretrained}.
For the contrastive loss, we prepend a learnable [CLS] token for each input view $\bm{v}$ before sending it into the IC-encoder and use the encoded [CLS] token as a view-level representation for computing the contrastive loss.
We use MoCo-v2~\cite{ji2021improving} with an exponential moving average (EMA) transformer encoder and a memory buffer to compute the contrastive loss.
More implementation details and hyper-parameters can be found in the supplementary.\looseness=-1

\begin{table}
\caption{
Image captioning results on the test set of MS-COCO Karpathy split~\cite{karpathy2015deep}.
Since our \heav{} is trained from scratch, we separately compare methods with large-scale transformer pre-training on the top block, and those trained from scratch on the bottom block.
Our \heav{} outperforms previous trained-from-scratch methods by a large margin and achieves comparable or often better performance compared to methods with large-scale transformer pre-training.
}\label{table:coco}
\centering
\renewcommand{\arraystretch}{1.2}
\resizebox{0.95\linewidth}{!}{
\begin{tabular}{@{\extracolsep{0pt}}lccccc@{}}
\toprule
Method & Pretrain Data & B-4 & M & C & S \\
\midrule
VLP~\cite{zhou2020unified} & 3M & 39.5 & 29.3 & 129.3 & 23.2 \\
X-VLM~\cite{zeng2021multi} & 16M & 40.4 & - & 139.3 & - \\
Oscar~\cite{li2020oscar} & 6.5M & 40.5 & 29.7 & 137.6 & 22.8 \\
VinVL~\cite{zhang2021vinvl} & 8.8M & 40.9 & 30.9 & 140.4 & \textbf{25.1} \\
GIT~\cite{wang2022git} & 4M & \textbf{41.3} & 30.4 & 139.1 & 24.3 \\
ViTCap~\cite{fang2022injecting} & 4M & 41.2 & 30.1 & 138.1 & 24.1 \\
$\text{SimVLM}_\text{base}$~\cite{wang2022simvlm} & 1.8B & 39.0 & \textbf{32.9} & 134.8 & 24.0 \\
\midrule
SCST~\cite{rennie2017self} & None & 34.2 & 26.7 & 114.0 & - \\
Up-Down~\cite{anderson2018bottom} & None & 36.3 & 27.7 & 120.1 & 21.4 \\
AoANet~\cite{huang2019attention} & None & 38.9 & 29.2 & 129.8 & 22.4 \\
$\mathcal{M}^2$~\cite{cornia2020meshed} & None & 39.1 & 29.1 & 131.2 & 22.6 \\
CLIP-ViL~\cite{shen2021much} & None & 40.2 & 29.7 & 134.2 & 23.8 \\
X-LAN~\cite{pan2020x} & None & 39.5 & 29.5 & 132.0 & 23.4 \\
DLCT~\cite{luo2021dual} & None & 39.8 & 29.5 & 133.8 & 23.0 \\
Xmodal-Ctx~\cite{kuo2022pretrained} & None & 39.7 & 30.0 & 135.9 & 23.7 \\
\heav{} (ours) & None & \textbf{41.0} & \textbf{30.2} & \textbf{141.5} & \textbf{23.9} \\
\bottomrule
\end{tabular}
}
\end{table}

\subsection{Main Results}
In Table~\ref{table:coco}, we show the results of our \heav{} on the test set of MS-COCO Karpathy split~\cite{karpathy2015deep}.
It is worth noting that \heav{} is \textbf{not} pre-trained on external image-and-text corpus and instead is trained from scratch only on the MS-COCO image captioning dataset~\cite{chen2015microsoft}.
Therefore, for a fair comparison, we separate these two different settings in Table~\ref{table:coco}, where methods on the top block use pre-training while those on the bottom are trained from scratch.
In the same trained-from-scratch setting, our \heav{} outperforms previous state-of-art Xmodal-Ctx~\cite{kuo2022pretrained} by 5.6\% in CIDEr and 1.3\% in BLEU-4.
On the other hand, when compared with methods with transformer pre-training, our \heav{}, despite being only trained on MS-COCO, achieves comparable or often better performance.

\begin{table}
\caption{
Image captioning results on a smaller-scale Flickr30K Karpathy split~\cite{karpathy2015deep}.
We also demonstrate semi-supervised learning (SSL) for image captioning with labeled data from Fickr30K and unlabeled data from MS-COCO.
Compared to previous state of the arts, our \heav{} achieves significant performance improvement, which may come from the label efficiency of our method.
}
\centering
\renewcommand{\arraystretch}{1.2}
\resizebox{0.9\linewidth}{!}{
\begin{tabular}{@{\extracolsep{4pt}}lcccc@{}}
\toprule
Method & B-4 & M & C & S \\
\midrule
Show \& Tell~\cite{vinyals2015show} & 21.5 & 18.3 & 41.7 & 12.2\\
Show, Attend \& Tell~\cite{xu2015show} & 23.6 & 19.2 & 49.1 & 13.3 \\
Up-Down~\cite{anderson2018bottom} & 28.3 & 21.6 & 63.3 & 15.9 \\
$\mathcal{M}^2$~\cite{cornia2020meshed} & 29.8 & 22.4 & 68.4 & 16.2 \\
ORT~\cite{herdade2019image} & 30.1 & 22.8 & 68.8 & 16.9 \\
\midrule
\heav{} (ours) & \textbf{34.3} & 24.6 & 81.7 & 18.0 \\
\heav{} + SSL (ours) & \textbf{34.3} & \textbf{25.1} & \textbf{85.6} & \textbf{19.0} \\
\bottomrule
\end{tabular}
}
\label{table:flickr}
\end{table}

To further demonstrate the label efficiency of \heav{}, we train it on the Karpathy split~\cite{karpathy2015deep} of Flickr30K, which only has $\sim$0.15M training data (4x smaller than MS-COCO), and show the results in Table~\ref{table:flickr}.
The results of other methods are taken from \cite{stefanini2022show}.
We can see that our \heav{} outperforms previous state-of-art ORT 
substantially by 12.9\% in CIDEr and 4.3\% in BLEU-4.
The significant improvement may come from the label efficiency of our method, which is particularly important when trained on the smaller-scale dataset of Flickr30K.

\subsection{Ablations and Analyses}

The goal of this paper is to propose an \textit{efficient} and \textit{effective} way to leverage heterogeneous views.
Therefore, in this section we closely examine whether our proposed \heav{} achieves these goals in Section~\ref{sec:efficient} for efficiency and Section~\ref{sec:effective} for effectiveness.
Following the convention in \cite{huang2019attention,kuo2022pretrained,herdade2019image}, we only train the model with cross-entropy loss for all ablations and analyses.

\subsubsection{Computation, parameter, and label efficient} \label{sec:efficient}

In Section~\ref{sec:encoding}, we propose to regard input views $\bm{v}$ as augmentations of the input image and encode the views independently with a shared IC-encoder.
To demonstrate the efficiency of our method, in Table~\ref{table:efficiency}, we show the theoretical computation and parameter complexity as well as the actual training speed and trainable parameters.
Since we do not want to trade performance in pursuit of efficiency,
we also show in Table~\ref{table:efficiency} that our method achieves better performance despite being more efficient.

\textbf{Comparison with common \vcombination{} approaches.} 
As described in Section~\ref{sec:heterogeneous-view}, other common approaches include (1) using a separate model per view~\cite{rennie2017self,cornia2020meshed,huang2019attention,anderson2018bottom}, which trains $|V|$ IC models, one for each input view, and averages the predictions of word probability from each IC model as the final word prediction, and (2) concatenation of views~\cite{li2020oscar,kuo2022pretrained}, which concatenates all input views along the sequence dimension into a long single view, and feeds the concatenated input views into the IC model for caption generation.
In Table~\ref{table:efficiency},
using a separate model per view has to train $|V|$ models and thus is parameters inefficient ($\mathcal{O}(|V|)$).
Concatenation of views concatenates all views into one long single view and is computationally expensive ($\mathcal{O}(|V|^2)$).
Compared to these two approaches, our \heav{} has linear computation ($\mathcal{O}(|V|)$) and constant parameter ($\mathcal{O}(1)$) complexity, and performs consistently better across all metrics by large margins.

\begin{table*}[t]
\caption{
Comparison with other encoding strategies in terms of efficiency and performance.
We train the models on MS-COCO Karpathy split~\cite{karpathy2015deep} and report the validation results.
All models are trained with cross-entropy loss only (\textit{w/o} SCST fine-tuning) following the convention in \cite{huang2019attention,kuo2022pretrained,herdade2019image}.
Our strategy of shared encoder trained with an additional contrastive loss $\mathcal{L}_{con}$ achieves the best performance and is computation and parameter efficient.
Please see Section~\ref{sec:efficient} for more details of different conditions.
}\label{table:efficiency}
\centering
\renewcommand{\arraystretch}{1.2}
\resizebox{0.9\linewidth}{!}{
\begin{tabular}{@{\extracolsep{4pt}}lcccccccccc@{}}
\toprule
& & \multicolumn{2}{c}{\textbf{Computation}} &  \multicolumn{2}{c}{\textbf{Parameter}} & \multicolumn{4}{c}{\textbf{Performance}}\\
\cline{3-4} \cline{5-6} \cline{7-10}
Conditions & $\mathcal{L}_\text{con}$ & complexity & iter/sec $\uparrow$ & complexity & \#params $\downarrow$ & B-4 & M & C & S \\
\midrule
Model per View & & $\mathcal{O}(|V|)$ & 2.23 & $\mathcal{O}(|V|)$ & 52.5M & 38.5 & 28.6 & 121.1 & 21.3 \\
Concatenated views & & $\mathcal{O}(|V|^2)$ & 4.20 & $\mathcal{O}(1)$ & 13.1M & 38.5 & 28.7 & 122.8 & 21.7 \\
Unshared encoders & & $\mathcal{O}(|V|)$ & 5.33 & $\mathcal{O}(|V|)$ & 20.8M & 39.7 & 29.1 & 125.4 & 22.1 \\
Unshared encoders & $\checkmark$ & $\mathcal{O}(|V|)$ & 3.96 & $\mathcal{O}(|V|)$ & 22.7M & 39.7 & 29.3 & 125.8 & 22.2 \\
Shared encoder & & $\mathcal{O}(|V|)$ & 5.97 & $\mathcal{O}(1)$ & 13.5M & 39.7 & 29.1 & 125.6 & 22.1 \\
\midrule
Ours (Shared encoder) & $\checkmark$ & $\mathcal{O}(|V|)$ & 4.83 & $\mathcal{O}(1)$ & 15.4M & \textbf{40.5} & \textbf{29.4} & \textbf{127.6} & \textbf{22.3} \\
\bottomrule
\end{tabular}
}
\end{table*}

\textbf{Shared \textit{v.s.} unshared encoder.} Another design choice is whether or not to use a shared encoder for each view.
In Table~\ref{table:efficiency}, in the case of training without a contrastive loss (\textit{w/o} $\mathcal{L}_{con}$), using unshared encoders~\cite{li2021align,akbari2021vatt,hu2021unit,alayrac2020self} for each view does not bring any performance benefits compared to using a shared encoder, but with the cost of higher parameter complexity ($\mathcal{O}(|V|)$).
Furthermore, we ablate the proposed contrastive loss $\mathcal{L}_{con}$, and found that the unshared encoder does not benefit from $\mathcal{L}_{con}$ as much as the shared encoder.
Although $\mathcal{L}_{con}$ introduces additional overhead such as extra parameters from the projection heads and extra computation from pairwise similarity during the training time (not for inference), the increase in complexity is moderate but the improvement in performance is significant when incorporated into our shared-encoder strategy.

\begin{figure}
\begin{center}
\includegraphics[width=0.85\linewidth]{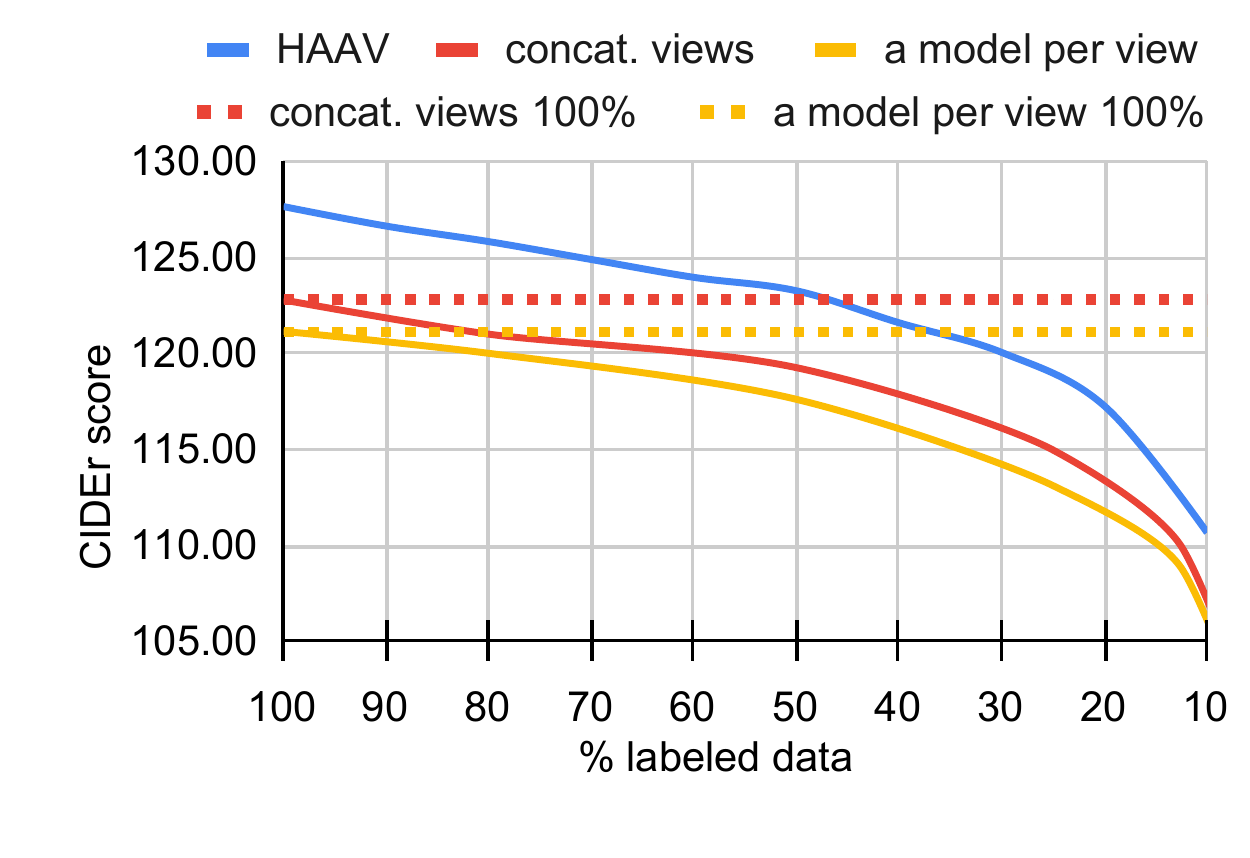}
\end{center}
\caption{
Data efficiency of \heav{}. \heav{} (\textcolor{blue}{blue}) only needs about 40-50\% labeled data to achieve similar performance as other common methods for heterogeneous view image captioning trained on 100\% of data such as using concatenated views (\textcolor{red}{red}) and a model per view (\textcolor{gold}{gold}).
All models are trained with cross-entropy loss only.\looseness=-1
}
\label{fig:data-efficiency}
\end{figure}

\textbf{Label efficiency with fewer training data.} To test the label efficiency of \heav{}, in Figure~\ref{fig:data-efficiency} we train the model with $\{10, 20,...,90\} \%$ of labeled data, and compare the CIDEr score with other common approaches for heterogeneous view image captioning, using concatenation of views (\textcolor{red}{red}) and a separate model per view (\textcolor{gold}{gold}), trained on 100\% of data.
We can see that \heav{} only needs about 40-50\% labeled data to achieve comparable performance.
With the same input views and similar model architectures, this indicates that our method is more label efficient, likely due to our novel use of heterogeneous views as augmentations and the contrastive loss $\mathcal{L}_{con}$.
In the supplementary, we also show that \heav{} suffers less from overfitting on MS-COCO in the supplementary, indicating that our model is more label efficient.\looseness=-1

\textbf{Semi-supervised training.} To demonstrate label efficiency brought by the contrastive loss $\mathcal{L}_{con}$ even further, we train a semi-supervised image captioning model.
Specifically, the model is trained on the labeled data (image-caption pairs) from Flickr30K, and unlabeled data (image only) from MS-COCO.
Since our contrastive loss is applied differently than other VL methods and does not require annotated pairs (\textit{e.g.} human-annotated image-caption pairs in MS-COCO), it can be applied on the unlabeled image-only data to aid representation learning of encoded views $\bm{u}$.
In Table~\ref{table:flickr}, with semi-supervised training, the already strong \heav{} achieves an even better performance of +3.9\% CIDEr and +1\% SPICE.\looseness=-1

\subsubsection{Effectiveness of hierarchical decoder layer}\label{sec:effective}

In Section~\ref{sec:decoding}, we propose a hierarchical decoder layer to model the effectiveness of encoded views and adaptively weigh them according to their effectiveness for caption generation.
To verify if the proposed hierarchical decoder achieves this goal, we monitor the view-level attention weights of $\text{CrossAttn}_{\text{Lv2}}$ in two control studies.
To show the architectural design of the proposed hierarchical decoder layer is effective, we compare with other common designs in Table~\ref{table:dec-arch}.

\begin{figure*}[t]
\centering
\begin{subfigure}{.37\linewidth}
  \centering
  \includegraphics[width=1.0\linewidth]{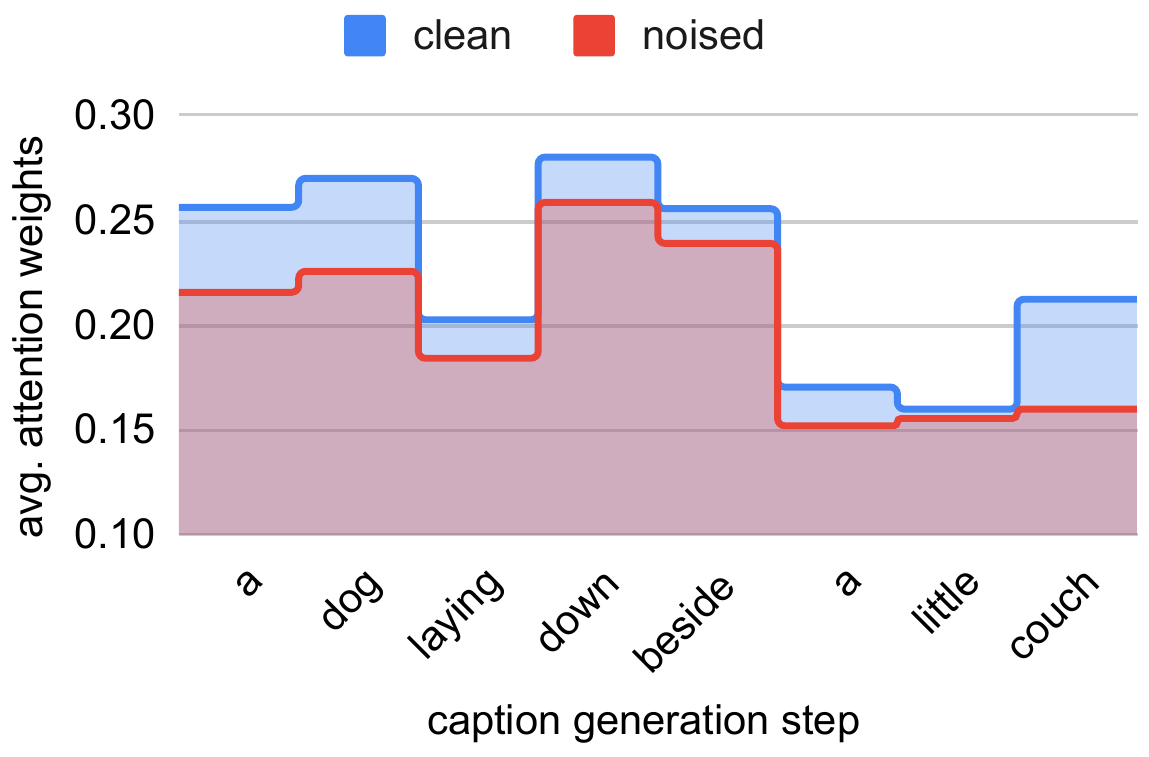}
  \caption{Attention weights averaged across heads for a noised view at each caption generation step.}\label{fig:beta-noised}
\end{subfigure}
\hfill
\begin{subfigure}{.25\linewidth}
  \centering
  \includegraphics[width=1.0\linewidth]{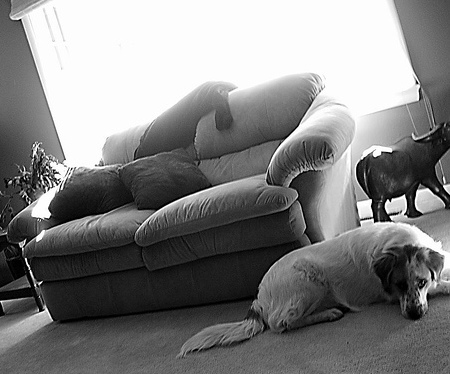}
  \begin{minipage}{.1cm}
    \vfill
  \end{minipage}
  \caption{Input image with caption: \textit{a dog laying down beside a little couch}}\label{fig:beta-image}
\end{subfigure}
\hfill
\begin{subfigure}{.35\linewidth}
  \centering
  \includegraphics[width=1.0\linewidth]{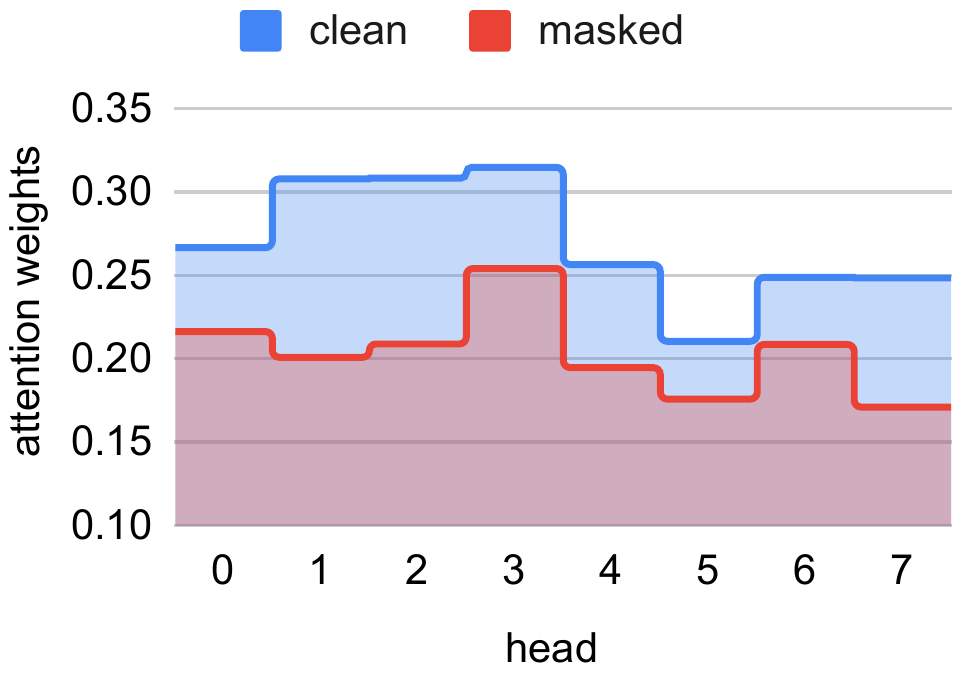}
  \caption{Attention weights of different attention heads at the step of generating ``\textit{dog}'', which is masked out in the input image.}\label{fig:beta-masked}
\end{subfigure}
\caption{
The view-level attention weights of $\text{CrossAttn}_{Lv2}$ vary adaptively according to the effectiveness of a view at the view level and at the word level when generating a caption for the center image.
\textbf{\textit{(a)}} We add random noise to a view and show that the attention weights averaged across different attention heads drop consistently at each step of caption generation.
\textbf{\textit{(c)}} We mask out dog in the input image and show that the attention weights of different attention heads drop consistently at the step of generating the word ``\textit{dog}''.
}\label{fig:beta}
\end{figure*}

\textbf{Adaptive view-level attention weights.}
We design two control studies in Figure~\ref{fig:beta} to show how the view-level attention weights of $\text{CrossAttn}_{Lv2}$ vary adaptively according to the effectiveness of an input view.
In the first experiment, we add noise to a view by randomly zeroing out tokens in a view to make a view \textit{less effective}, and expect a drop of weights toward that noised view.
To measure the weights, we take the multi-head attention weights of $\text{CrossAttn}_\text{Lv2}$ at the last decoder layer and average the attention weights across heads.
In Figure~\ref{fig:beta-noised}, the weights for the noised view drops consistently at each word prediction step compared to the same view \textit{without} added noise.
This means that our hierarchical decoder indeed learns to adaptively weigh the views according to their effectiveness at the view level.
In the second experiment, we randomly mask out a prominent region of the input image for a view.
For example, we mask out dog in the input image (Figure~\ref{fig:beta-image}) with caption ``\textit{a \textbf{dog} laying down beside a little couch}'' to make a view \textit{less effective} at the step of generating the word ``\textit{dog}''.
We expect a drop of the weights toward the masked view at the step of generating the word ``\textit{dog}''.
To measure the weights, we take the multi-head attention weights of $\text{CrossAttn}_\text{Lv2}$ at the last decoder layer and measure the attention weights of each head at the step of generating the word ``\textit{dog}''.
In Figure~\ref{fig:beta-masked}, the weights for the masked view drops consistently across all attention heads compared to the same view without masking.
This means that our hierarchical decoder indeed learns to adaptively weigh the input views according to their usefulness at the word level.
The example in Figure~\ref{fig:beta} is randomly chosen and more examples can be found in the supplementary.

\begin{table}
\caption{
Ablations of design choices for the proposed hierarchical decoder layer.
Our proposed two-tiered cross-attention architecture works the best.
The models are trained with cross-entropy only.
}\label{table:dec-arch}
\centering
\renewcommand{\arraystretch}{1.2}
\resizebox{0.75\linewidth}{!}{
\begin{tabular}{@{\extracolsep{4pt}}lcccc@{}}
\toprule
Method & B-4 & M & C & S \\
\midrule
Concatenate & 39.2 & 29.2 &  125.1 & 22.1 \\
Mean pool & 39.4 & 29.2 & 125.2 & 22.2 \\
Max pool & 38.7 & 28.8 & 122.9 & 21.5 \\
Sigmoid & 39.4 & 29.2 & 125.5 & 22.2 \\
Tanh & 39.4 & 29.3 & 124.8 & 22.0 \\
\midrule
Ours & \textbf{40.5} & \textbf{29.4} & \textbf{127.6} & \textbf{22.3} \\
\bottomrule
\end{tabular}
}
\end{table}

\textbf{Design choices of the hierarchical decoder.}
In Table~\ref{table:dec-arch}, we ablate different design choices of the hierarchical decoder.
One simple alternative is to concatenate the encoded views $\bm{u}$ along the sequence dimension into a long single view, and use a single cross-attention module to jointly \vcombine{} the views at the token level.
One can also mean-/max-pool to \vcombine{} across views in place of the $\text{CrossAttn}_\text{Lv2}$ module.
Other works also propose to use a sigmoid/tanh gating mechanism~\cite{huang2019attention} to \vcombine{} across layers~\cite{cornia2020meshed}, which can be generalized to \vcombining{} across views.
In Table~\ref{table:dec-arch}, we can see that our proposed hierarchical decoder layer with a two-tiered cross-attention structure achieves the best performance compared to all other designs.\looseness=-1
\section{Conclusion}
In this paper, we focus on the problem of how to \textit{efficiently} and \textit{effectively} leverage heterogeneous views $\bm{v}$.
To tackle this problem, we propose \heav{} to (1) regard heterogeneous views as augmentations of the input image, and naturally encode each view independently with a shared encoder in an efficient manner, (2) incorporate a contrastive loss across encoded views $\bm{u}$ to improve representation quality and label efficiency (especially  semi-supervised training to leverage image-only unlabeled data), and (3) carefully design a hierarchical decoder layer that first \vcombines{} within each view at the token level and then across views at the view level to account for the effectiveness of each encoded view $\bm{u}$.
Through rigorous analysis, we show \heav{} is computation, parameter, and data efficient, and outperforms other less efficient designs and existing approaches for heterogeneous views image captioning.
We also demonstrate that our hierarchical decoder successfully models the effectiveness of views and weigh them adaptively according to their effectiveness.
We demonstrate significant performance improvements of +5.6\% CIDEr compared to state-of-art \textit{w/o} transformer pre-training on MS-COCO and +16.8\% CIDEr with SSL on Flickr30K.
\section*{Acknowledgement}

This work was funded by DARPA's Learning with Less Labels (LwLL) program under agreement HR0011-18-S-0044

{\small
\bibliographystyle{ieee_fullname}
\bibliography{section/egbib}
}
\clearpage

\appendix

\section*{Appendix}

\section{COCO test results}

\begin{table*}[b!]
\caption{MS-COCO test server results}\label{table:coco-test}
\vspace{-7pt}
\centering
\renewcommand{\arraystretch}{1.2}
\resizebox{0.85\linewidth}{!}{
\begin{tabular}{@{\extracolsep{-2pt}}ccccccccccccccc@{}}
\toprule
&\multicolumn{2}{c}{\textbf{B-1}} & \multicolumn{2}{c}{\textbf{B-2}} & \multicolumn{2}{c}{\textbf{B-3}} & \multicolumn{2}{c}{\textbf{B-4}} & \multicolumn{2}{c}{\textbf{M}} & \multicolumn{2}{c}{\textbf{R}} & \multicolumn{2}{c}{\textbf{C}}\\
\cline{2-3} \cline{4-5} \cline{6-7} \cline{8-9} \cline{10-11} \cline{12-13} \cline{14-15}
Method & c5 & c40 & c5 & c40 & c5 & c40 & c5 & c40 & c5 & c40 & c5 & c40 & c5 & c40\\
\midrule
SCST~\cite{rennie2017self} & 78.1 & 93.7 & 61.9 & 86.0 & 47.0 & 75.9 & 35.2 & 64.5 & 27.0 & 35.5 & 56.3 & 70.7 & 114.7 & 116.7 \\
Up-Down~\cite{anderson2018bottom} & 80.2 & 95.2 & 64.1 & 88.8 & 49.1 & 79.4 & 36.9 & 68.5 & 27.6 & 36.7 & 57.1 & 72.4 & 117.9 & 120.5 \\
AoANet~\cite{huang2019attention} & 81.0 & 95.0 & 65.8 & 89.6 & 51.4 & 81.3 & 39.4 & 71.2 & 29.1 & 38.5 & 58.9 & 74.5 & 126.9 & 129.6 \\
$\mathcal{M}^2$~\cite{cornia2020meshed} & 81.6 & 96.0 & 66.4 & 90.8 & 51.8 & 82.7 & 39.7 & 72.8 & 29.4 & 39.0 & 59.2 & 74.8 & 129.3 & 132.1 \\
X-LAN~\cite{pan2020x} & 81.9 & 95.7 & 66.9 & 90.5 & 52.4 & 82.5 & 40.3 & 72.4 & 29.6 & 39.2 & 59.5 & 75.0 & 131.1 & 133.5 \\
DLCT~\cite{luo2021dual} & 82.4 & 96.6 & 67.4 & 91.7 & 52.8 & 83.8 & 40.6 & 74.0 & 29.8 & 39.6 & 59.8 & 75.3 & 133.3 & 135.4 \\
HAAV (ours) & \textbf{84.0} & \textbf{97.6} & 69.1 & 93.3 & 54.3 & 85.8 & \textbf{41.7} & \textbf{76.1} & \textbf{30.2} & \textbf{39.9} & \textbf{60.4} & \textbf{75.8} & \textbf{139.1} & \textbf{142.3} \\
\bottomrule
\end{tabular}
}
\end{table*}
We also evaluate our HAAV using an ensemble of four models same as other methods on the online MS-COCO test server, and report the results in Table~\ref{table:coco-test}.
We can see that our HAAV outperforms previous methods by a large margin across all metrics.

\section{Overfitting}\label{sec:supp:overfitting}

\begin{figure}[t!]
\centering
\includegraphics[width=\linewidth]{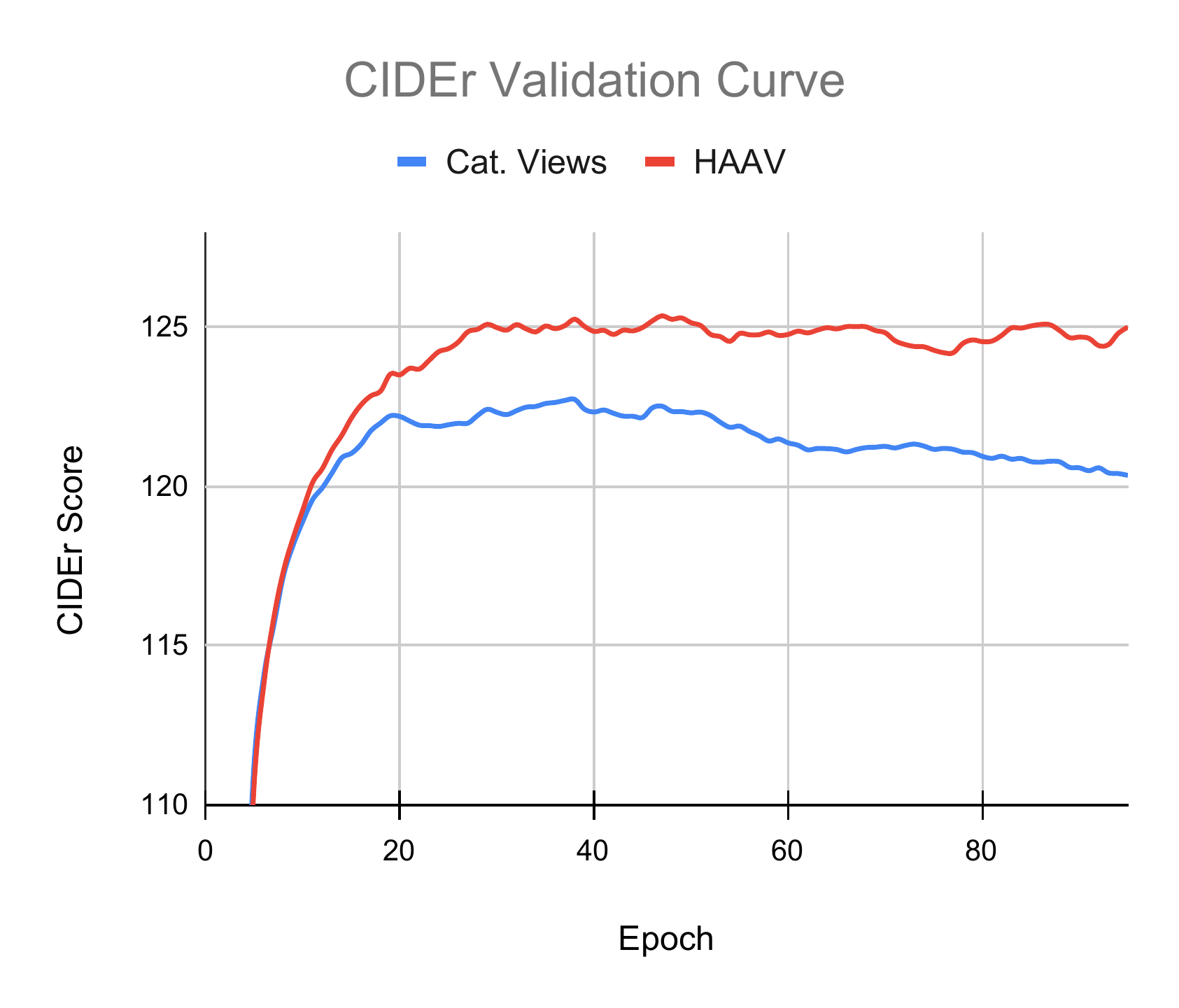}
\caption{
CIDEr validation curve of HAAV \textit{v.s.} concatenated views.
}\label{fig:val_curve}
\end{figure}

Training a data-hungry transformer model on a medium-scale dataset of MS-COCO (around 0.6M training samples) is prone to overfitting.
In HAAV, we propose to regard heterogeneous views as augmentations of the input image and encode the views independently with a shared encoder.
We claim that this formulation increases data diversity and is more parameter and label-efficient.
Furthermore, we add a contrastive loss to improve representation quality of encoded views, which is also beneficial for label efficiency.
In Figure~\ref{fig:val_curve}, we show the validation curve for concatenated views and HAAV. Compared to concatenated views, our HAAV indeed suffers less from overfitting.
Due to overfitting, the CIDEr score of concatenated views drops by 3.6 from the highest to the end of training.

\section{Implementation Details}\label{sec:supp:impl}
\begin{table*}[h!]
\caption{
Hyperparameters for cross-entropy training.
The values for untuned parameters are inherited from the base image captioning model, Xmodal-Ctx~\cite{kuo2022pretrained}.
}\label{table:param-xe}
\centering
\renewcommand{\arraystretch}{1.2}
\resizebox{0.7\linewidth}{!}{
\begin{tabular}{@{\extracolsep{4pt}}lccl@{}}
\toprule
Hyperparameter & Value & Tuned & Note \\
\midrule
N & 3 & & Number of encoder layers \\
M & 3 & & Number of decoder layers \\
lr & 2e-5 & \checkmark & learning rate \\
bs & 50 & & batch size \\
wd & 0.05 & & weight decay \\
$\lambda$ & 0.05 & \checkmark & loss weight for $\mathcal{L}_{con}$ \\
$p_c$ & 0.1 & & drop rate for channel-wise dropout \\
$p_s$ & 0.1 & & drop rate for sequence-wise dropout \\
$p_v$ & 0.1 & & drop rate for view-wise dropout \\
optimizer & AdamW & & Adam with decoupled weight decay ~\cite{Loshchilov2019DecoupledWD} \\ 
lr scheduler & \makecell{constant with \\ linear warmup} & & \makecell[l]{linearly warm up lr from 0.0,\\ and then stay constant}\\
warmup steps & 10k \\
K & 8k & \checkmark & \makecell[l]{size of memory buffer\\ for MoCo contrastive learning} \\
$\tau$ & 0.06 & \checkmark & \makecell[l]{temperature scaling\\ for MoCo contrastive learning} \\
ema & 0.999 & & \makecell[l]{exponential moving avergae\\ for MoCo contrastive learning} \\
\bottomrule
\end{tabular}
}
\end{table*}

\begin{table*}[h!]
\caption{
Hyperparameters for SCST~\cite{rennie2017self} training.
The values for untuned parameters are inherited from the base image captioning model, Xmodal-Ctx~\cite{kuo2022pretrained}, or from the tuned value in cross-entropy training.
}\label{table:param-scst}
\centering
\renewcommand{\arraystretch}{1.2}
\resizebox{0.7\linewidth}{!}{
\begin{tabular}{@{\extracolsep{4pt}}lccl@{}}
\toprule
Hyperparameter & Value & Tuned & Note \\
\midrule
N & 3 & & Number of encoder layers \\
M & 3 & & Number of decoder layers \\
lr & 5e-6 & & learning rate \\
bs & 40 & & batch size \\
wd & 0.0 & & weight decay \\
$\lambda$ & 0.2 & \checkmark & loss weight for $\mathcal{L}_{con}$ \\
$p_c$ & 0.1 & & drop rate for channel-wise dropout \\
$p_s$ & 0.1 & & drop rate for sequence-wise dropout \\
$p_v$ & 0.1 & & drop rate for view-wise dropout \\
optimizer & AdamW & & Adam with decoupled weight decay~\cite{Loshchilov2019DecoupledWD} \\ 
lr scheduler & None & & do not use any lr scheduler\\
K & 8k & \checkmark & \makecell[l]{size of memory buffer\\ for MoCo contrastive learning} \\
$\tau$ & 0.06 & \checkmark & \makecell[l]{temperature scaling\\ for MoCo contrastive learning} \\
ema & 0.999 & & \makecell[l]{exponential moving avergae\\ for MoCo contrastive learning}\\
\bottomrule
\end{tabular}
}
\end{table*}

We provide a detailed list of hyperparameters including their values and whether they are tuned in Table~\ref{table:param-xe} (cross-entropy training) and Table~\ref{table:param-scst} (SCST training).
For cross-entropy training, the model can be trained with a single Nvidia 2080 Ti GPU in 2 days.
For SCST training, the model can be trained with a single Nvidia A40 GPUs in 4 days.\looseness=-1

~\newpage~\newpage
\section{Generated Captions}

\begin{figure*}[b!]
\centering
\begin{subfigure}[t]{.27\linewidth}
  \centering
  \includegraphics[width=0.56\linewidth]{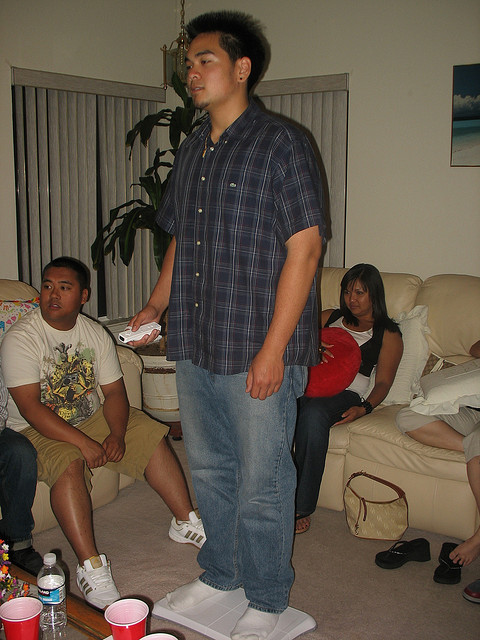}
  \caption{\\
  \textbf{HAAV}: a man standing in a living room holding a nintendo wii game controller \\
  \textbf{Xmodal-Ctx}: a group of people sitting on a couch playing a video game
  }
  \label{fig:caption-a}
\end{subfigure}\hfill
\begin{subfigure}[t]{.23\linewidth}
  \centering
  \includegraphics[width=1.0\linewidth]{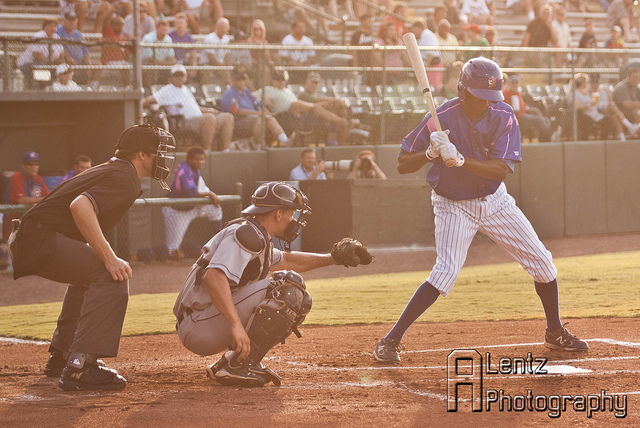}
  \caption{\\
  \textbf{HAAV}: a batter catcher and umpire during a baseball game \\
  \textbf{Xmodal-Ctx}: a baseball player holding a bat on a field \\ \\ 
  }
\end{subfigure}\hfill
\begin{subfigure}[t]{.23\linewidth}
  \centering
  \includegraphics[width=1.0\linewidth]{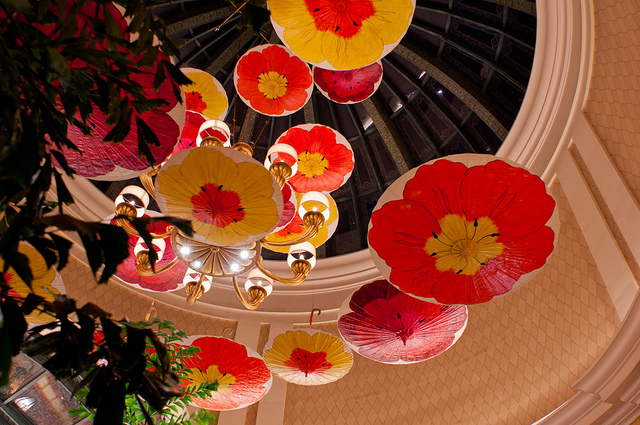}
  \caption{\\
  \textbf{HAAV}: a bunch of umbrellas hanging from a ceiling \\
  \textbf{Xmodal-Ctx}: a bunch of flowers hanging from a ceiling \\ \\
  }
\end{subfigure}\hfill
\begin{subfigure}[t]{.23\linewidth}
  \centering
  \includegraphics[width=1.0\linewidth]{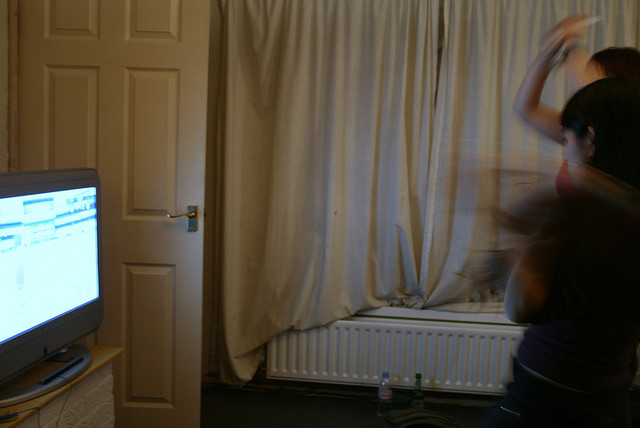}
  \caption{\\
  \textbf{HAAV}: two people playing a video game in a room \\
  \textbf{Xmodal-Ctx}: a person standing in front of a tv
  }
\end{subfigure}\hfill
\begin{subfigure}[t]{.24\linewidth}
  \centering
  \includegraphics[width=1.0\linewidth]{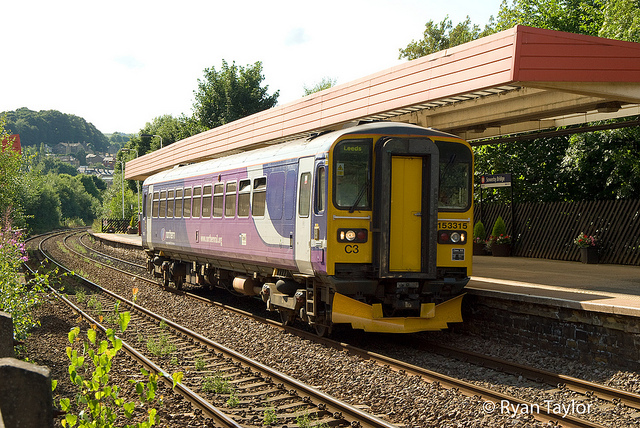}
  \caption{\\
  \textbf{HAAV}: a yellow and purple train parked at a train station \\
  \textbf{Xmodal-Ctx}: a train that is sitting on the tracks
  }\label{fig:caption-e}
\end{subfigure}\hfill
\begin{subfigure}[t]{.24\linewidth}
  \centering
  \includegraphics[width=1.0\linewidth]{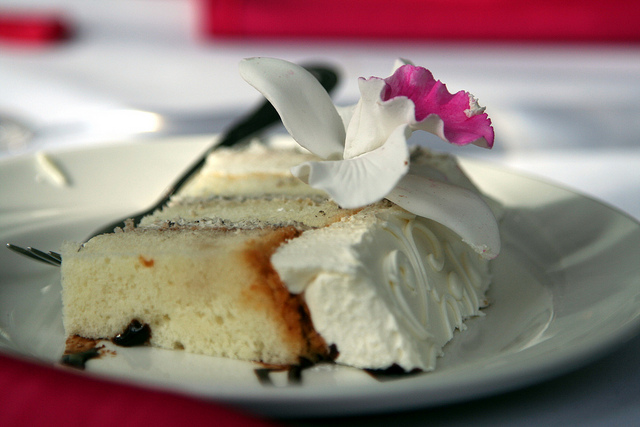}
  \caption{\\
  \textbf{HAAV}: a piece of cake on a plate with a flower \\
  \textbf{Xmodal-Ctx}: a slice of cake on a plate with a fork
  }
\end{subfigure}\hfill
\begin{subfigure}[t]{.24\linewidth}
  \centering
  \includegraphics[width=1.0\linewidth]{figure/COCO_val2014_000000262197.jpg}
  \caption{\\
  \textbf{HAAV}: a building with a clock tower in the middle of it \\
  \textbf{Xmodal-Ctx}: a large building with a clock on the side of it
  }
\end{subfigure}\hfill
\begin{subfigure}[t]{.24\linewidth}
  \centering
  \includegraphics[width=1.0\linewidth]{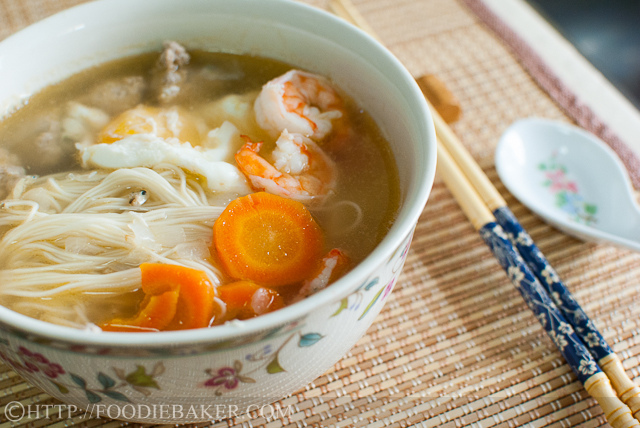}
  \caption{\\
  \textbf{HAAV}: a bowl of soup with vegetables and noodles \\
  \textbf{Xmodal-Ctx}: a bowl of soup sitting on top of a table
  }\label{fig:caption-h}
\end{subfigure}\hfill
\caption{
Captions generated by HAAV and another trained-from-scratch SoTA method Xmodal-Ctx~\cite{kuo2022pretrained}
}\label{fig:generated-captions}
\end{figure*}

In Figure~\ref{fig:generated-captions}, we show some random examples of different captions generated by our HAAV and another trained-from-scratch SoTA method Xmodal-Ctx~\cite{kuo2022pretrained}.
Qualitatively, HAAV is capable of generating captions in more details and more closely related to the input image rather than generating a generic sentence.
For example, in Figure~\ref{fig:caption-a}, HAAV generates ``\textit{a man standing in a living room holding a nintendo wii game controller}'', while Xmodal-Ctx generates a more generic description of ``\textit{a group of people sitting on a couch playing a video game}''.
Another example in Figure~\ref{fig:caption-e} shows that HAAV describes the train in more details as ``\textit{a yellow and purple train}'' rather than just ``\textit{a train}'' by Xmodal-Ctx.

\section{Adaptive View Aggregation Weights}\label{sec:supp:beta}

\begin{figure*}[h!]
\centering
\begin{subfigure}{.39\linewidth}
  \centering
  \includegraphics[width=1.0\linewidth]{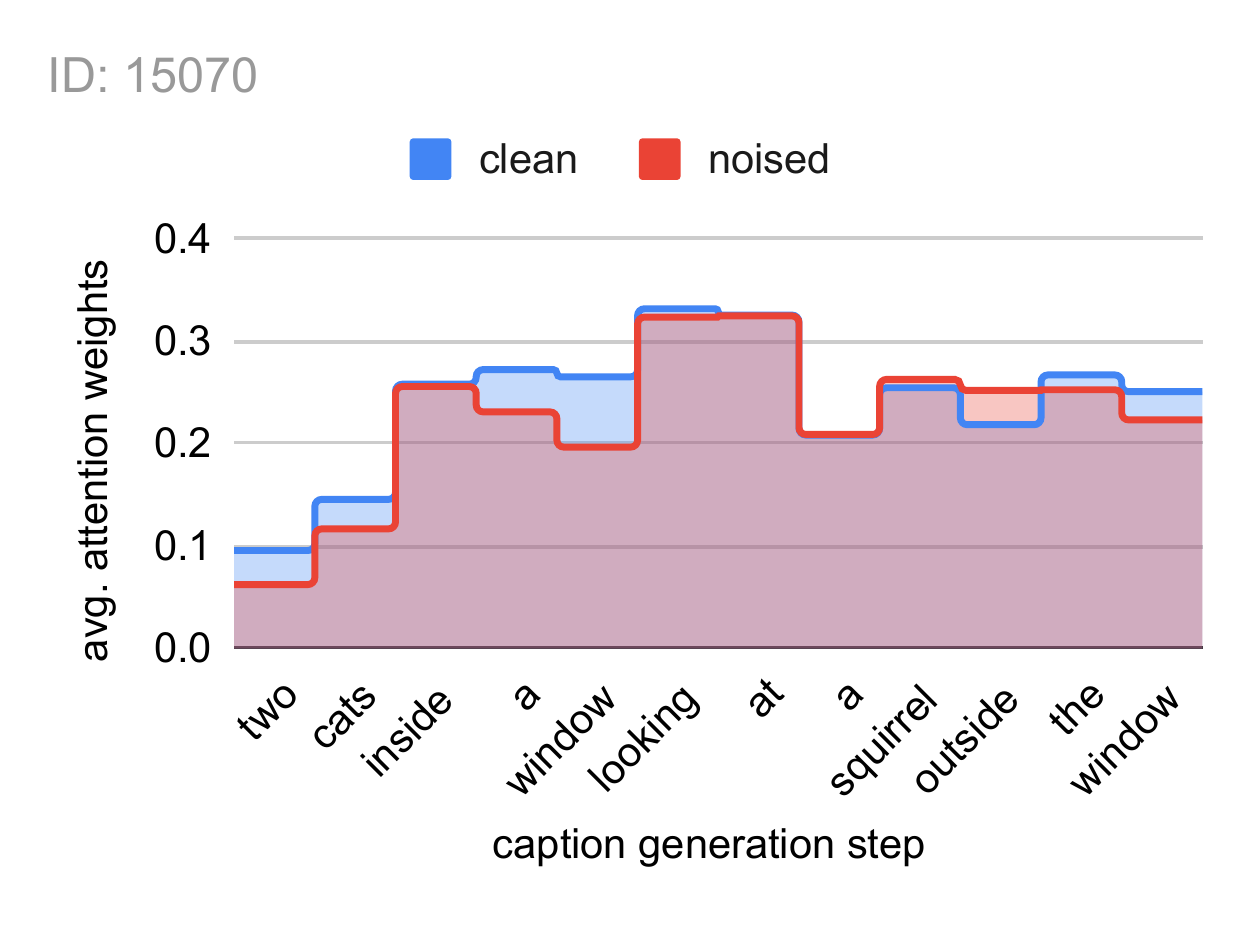}
\end{subfigure}\hfill
\begin{subfigure}{.25\linewidth}
  \centering
  \includegraphics[width=1.0\linewidth]{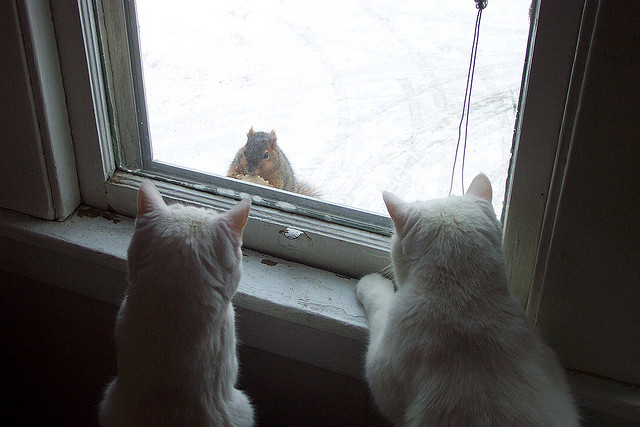}
  \begin{minipage}{.1cm}
    \vfill
  \end{minipage}
\end{subfigure}\hfill
\begin{subfigure}{.34\linewidth}
  \centering
  \includegraphics[width=1.0\linewidth]{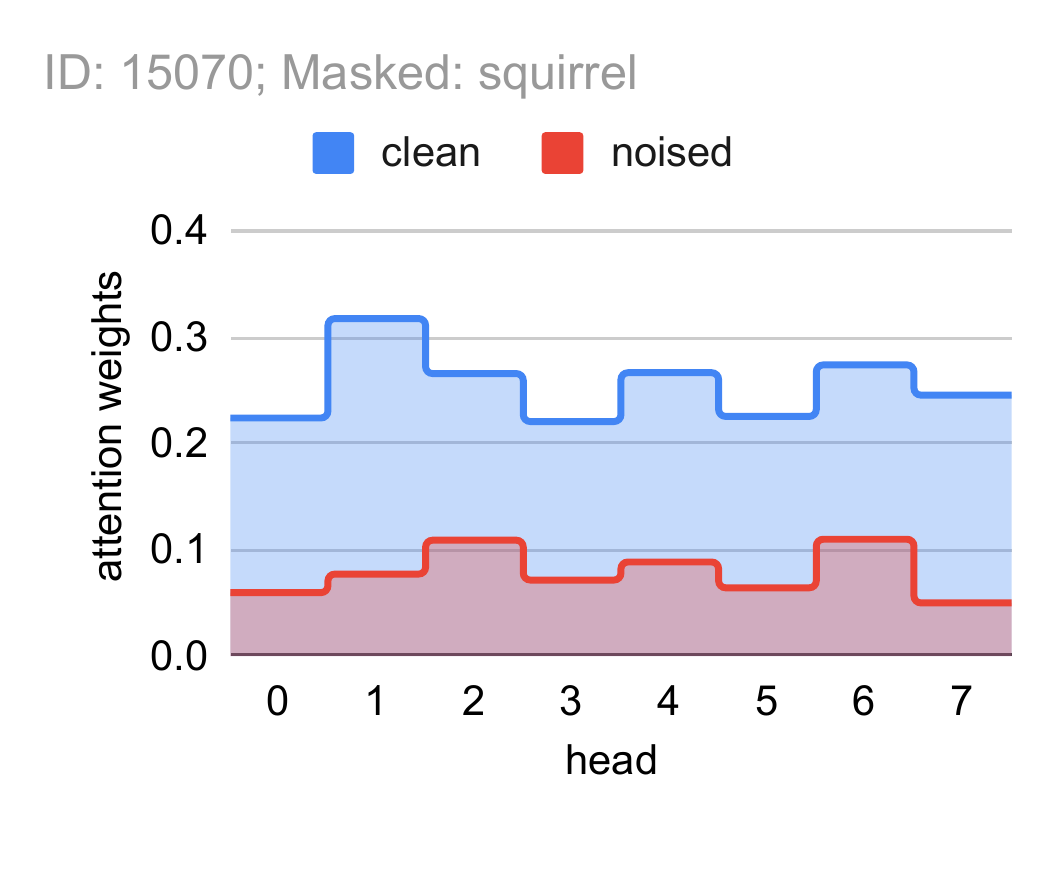}
\end{subfigure}
\vspace{-15pt}
\caption{
\textbf{\textit{(left)}} The attention weights averaged across different attention heads for a noised view drop consistently at each caption generation step.
\textbf{\textit{(center)}} Input image with caption: ``\textit{two cats inside a window looking at a squirrel outside the window}''.
\textbf{\textit{(right)}} The attention weights of different attention heads drop consistently at the step of generating the word ``\textit{squirrel}'', which is masked out in the input image.
}\label{fig:supp:beta1}
\end{figure*}

\begin{figure*}[h!]
\centering
\begin{subfigure}{.39\linewidth}
  \centering
  \includegraphics[width=1.0\linewidth]{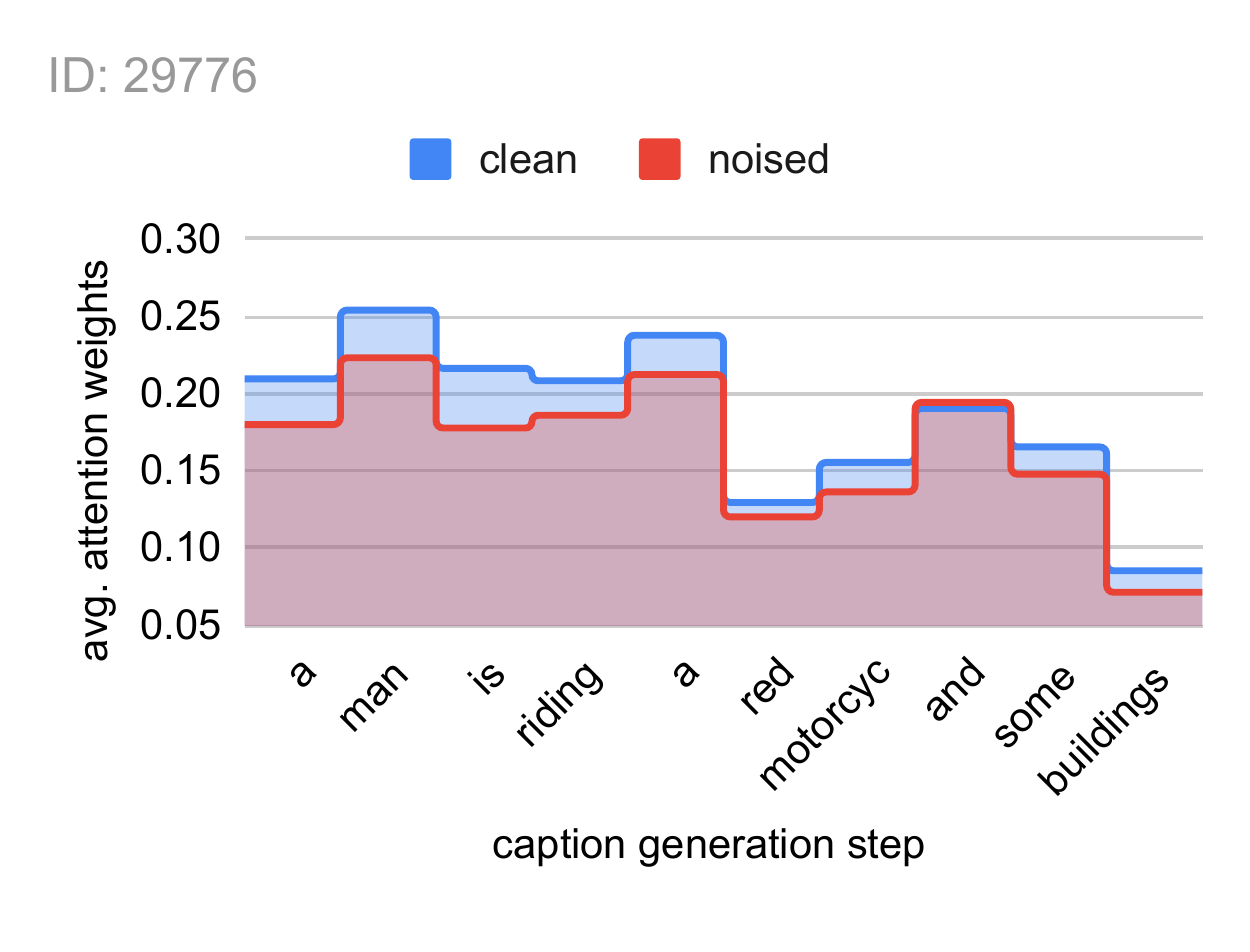}
\end{subfigure}\hfill
\begin{subfigure}{.25\linewidth}
  \centering
  \includegraphics[width=1.0\linewidth]{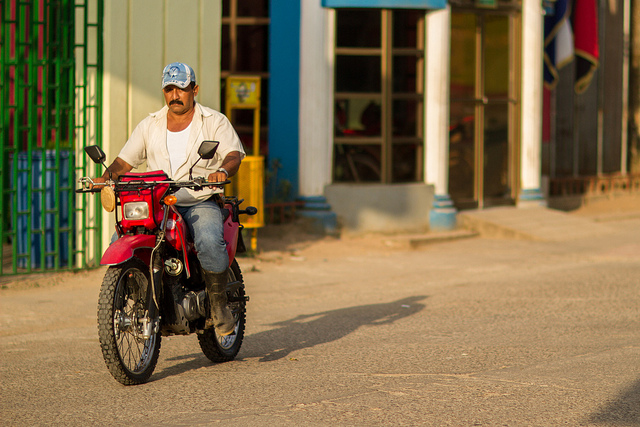}
  \begin{minipage}{.1cm}
    \vfill
  \end{minipage}
\end{subfigure}\hfill
\begin{subfigure}{.34\linewidth}
  \centering
  \includegraphics[width=1.0\linewidth]{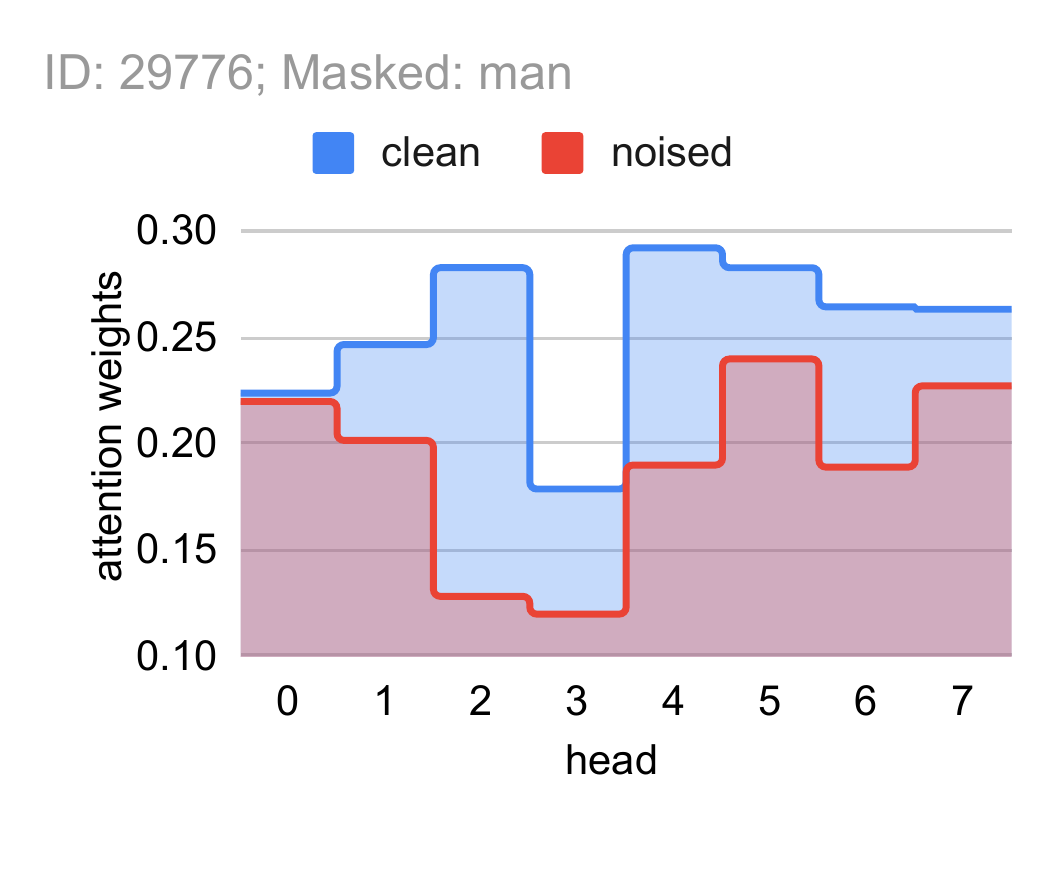}
\end{subfigure}
\vspace{-15pt}
\caption{
\textbf{\textit{(left)}} The attention weights averaged across different attention heads for a noised view drop consistently at each caption generation step.
\textbf{\textit{(center)}} Input image with caption: ``\textit{a man is riding a red motorcycle and some buildings}''.
\textbf{\textit{(right)}} The attention weights of different attention heads drop consistently at the step of generating the word ``\textit{man}'', which is masked out in the input image.
}\label{fig:supp:beta2}
\end{figure*}

\begin{figure*}[h!]
\centering
\begin{subfigure}{.39\linewidth}
  \centering
  \includegraphics[width=1.0\linewidth]{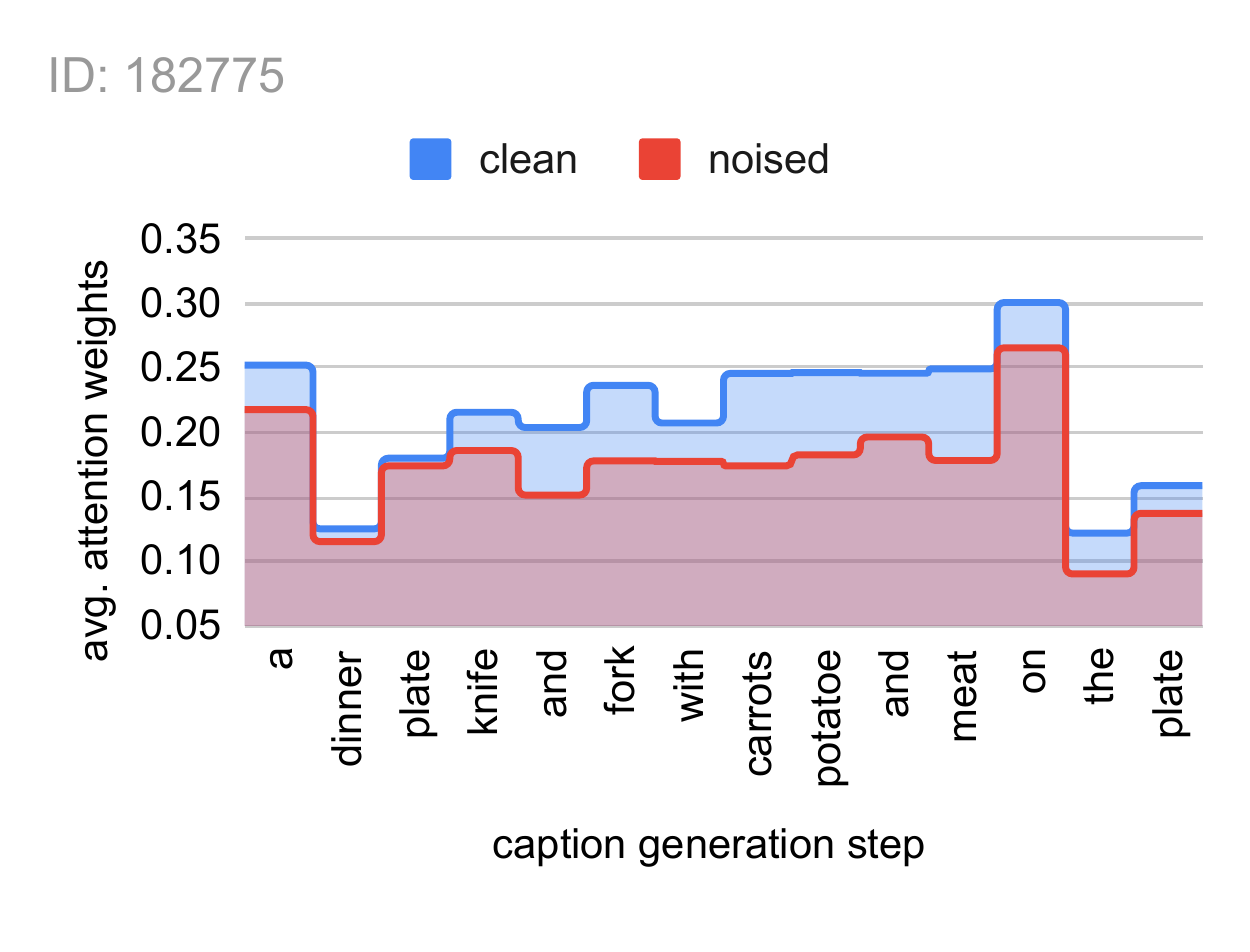}
\end{subfigure}\hfill
\begin{subfigure}{.25\linewidth}
  \centering
  \includegraphics[width=1.0\linewidth]{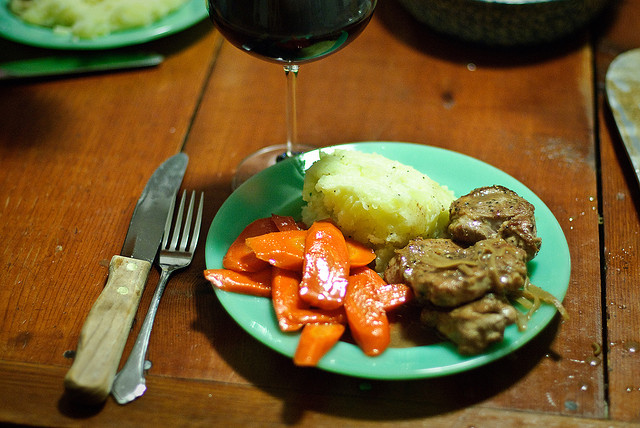}
  \begin{minipage}{.1cm}
    \vfill
  \end{minipage}
\end{subfigure}\hfill
\begin{subfigure}{.34\linewidth}
  \centering
  \includegraphics[width=1.0\linewidth]{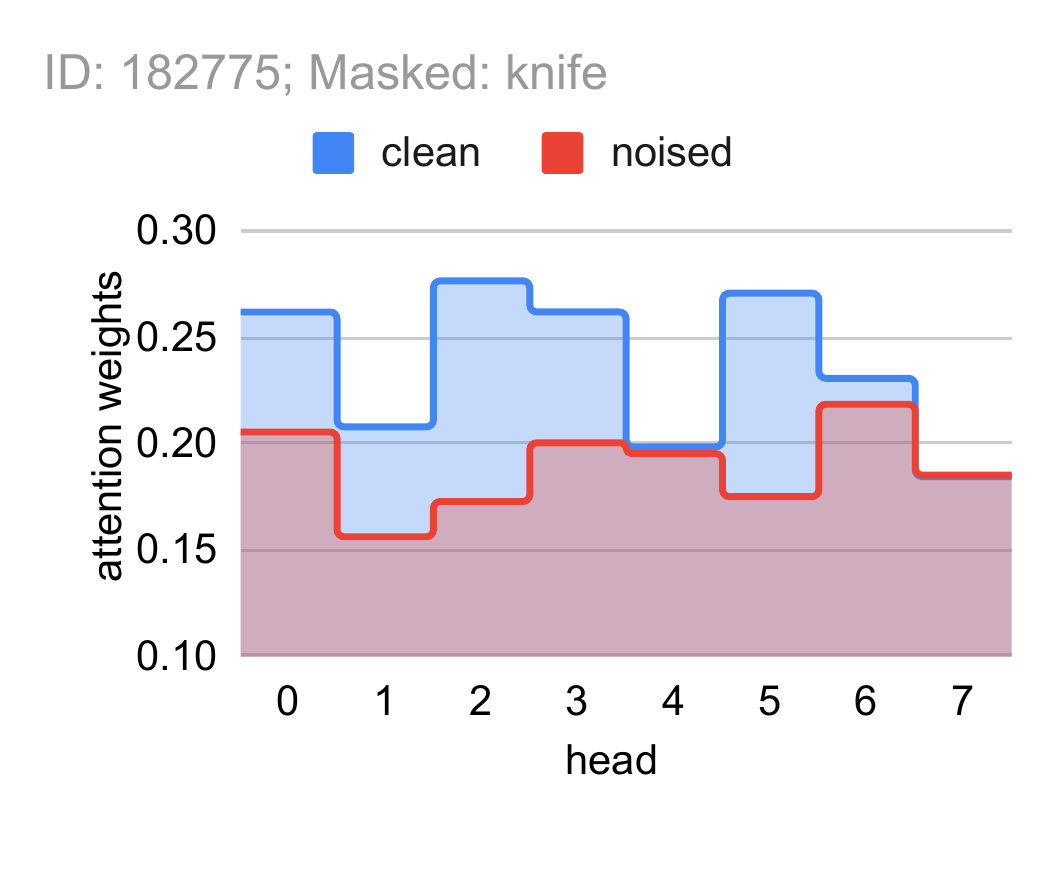}
\end{subfigure}
\vspace{-15pt}
\caption{
\textbf{\textit{(left)}} The attention weights averaged across different attention heads for a noised view drop consistently at each caption generation step.
\textbf{\textit{(center)}} Input image with caption: ``\textit{a dinner plate knife and fork with carrots potatoes and meat on the plate}''.
\textbf{\textit{(right)}} The attention weights of different attention heads drop consistently at the step of generating the word ``\textit{knife}'', which is masked out in the input image.
}\label{fig:supp:beta3}
\end{figure*}

\begin{figure*}[h!]
\centering
\begin{subfigure}{.39\linewidth}
  \centering
  \includegraphics[width=1.0\linewidth]{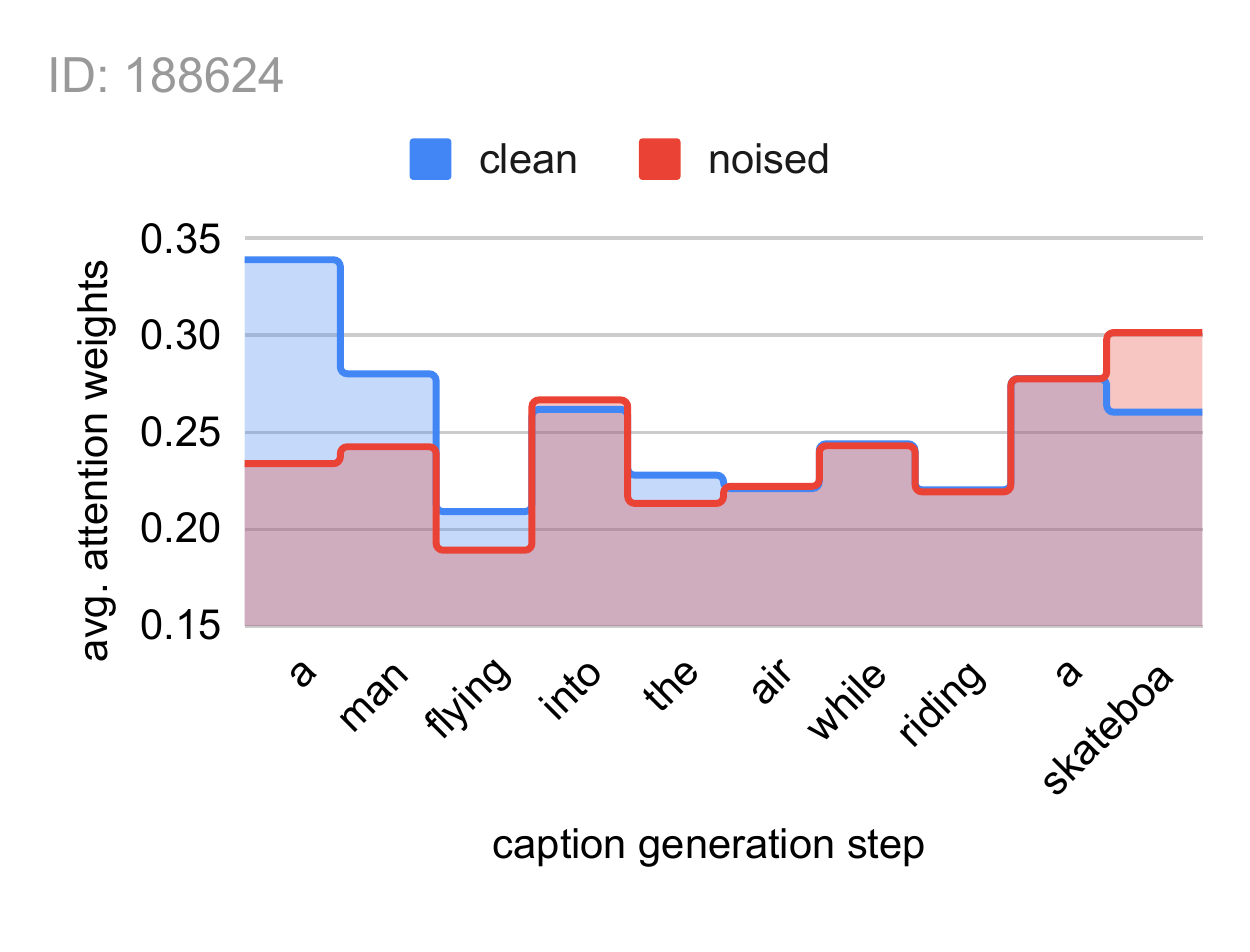}
\end{subfigure}\hfill
\begin{subfigure}{.25\linewidth}
  \centering
  \includegraphics[width=1.0\linewidth]{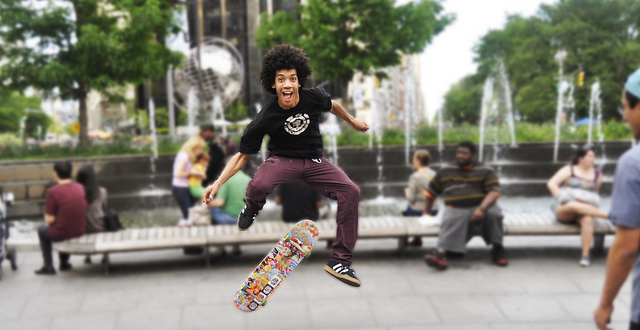}
  \begin{minipage}{.1cm}
    \vfill
  \end{minipage}
\end{subfigure}\hfill
\begin{subfigure}{.34\linewidth}
  \centering
  \includegraphics[width=1.0\linewidth]{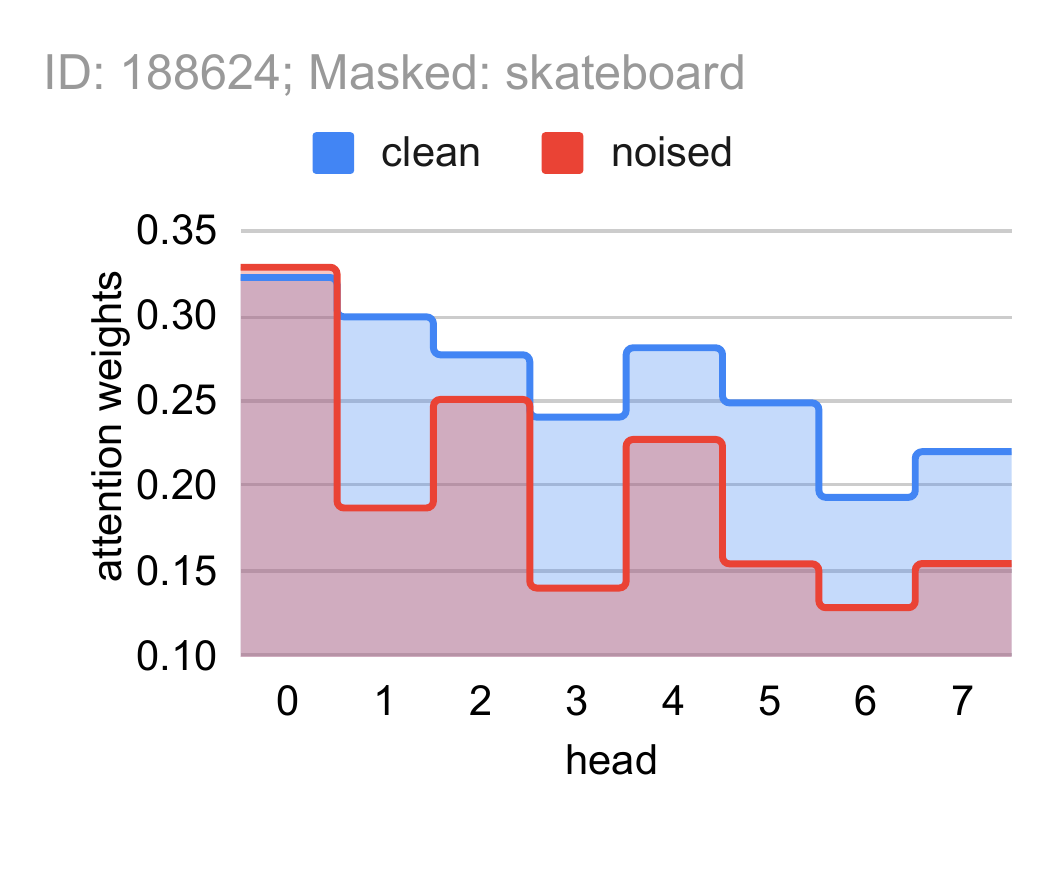}
\end{subfigure}
\vspace{-16pt}
\caption{
\textbf{\textit{(left)}} The attention weights averaged across different attention heads for a noised view drop consistently at each caption generation step.
\textbf{\textit{(center)}} Input image with caption: ``\textit{a man flying into the air while riding a skateboard}''.
\textbf{\textit{(right)}} The attention weights of different attention heads drop consistently at the step of generating the word ``\textit{skateboard}'', which is masked out in the input image.
}\label{fig:supp:beta4}
\end{figure*}

\begin{figure*}[h!]
\centering
\begin{subfigure}{.39\linewidth}
  \centering
  \includegraphics[width=1.0\linewidth]{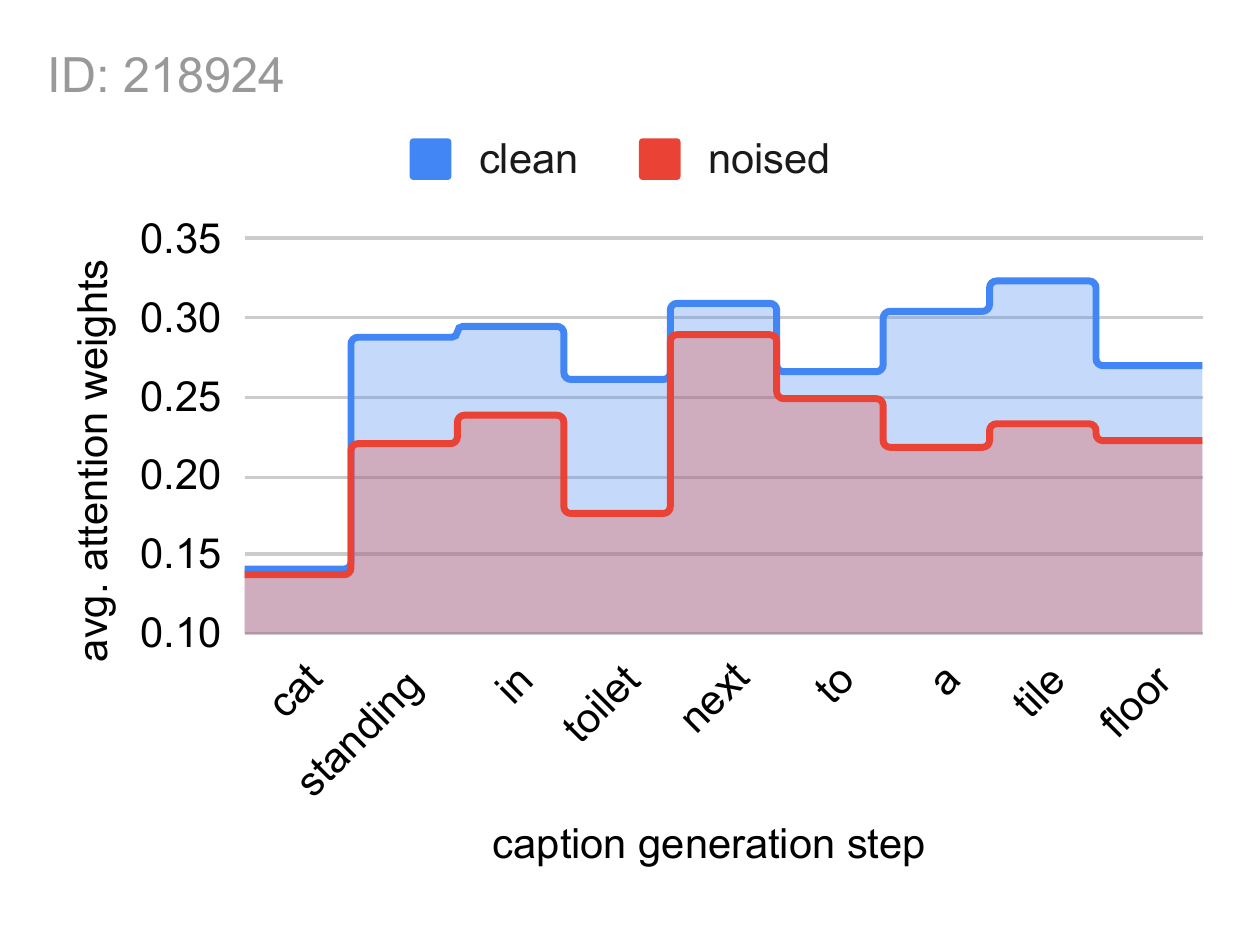}
\end{subfigure}\hfill
\begin{subfigure}{.25\linewidth}
  \centering
  \includegraphics[width=1.0\linewidth]{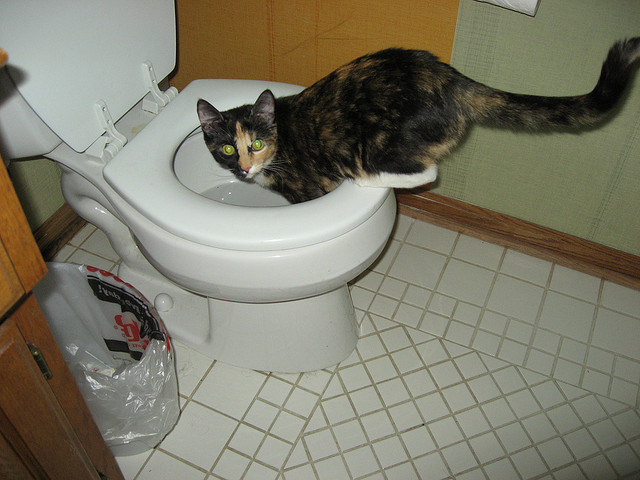}
  \begin{minipage}{.1cm}
    \vfill
  \end{minipage}
\end{subfigure}\hfill
\begin{subfigure}{.34\linewidth}
  \centering
  \includegraphics[width=1.0\linewidth]{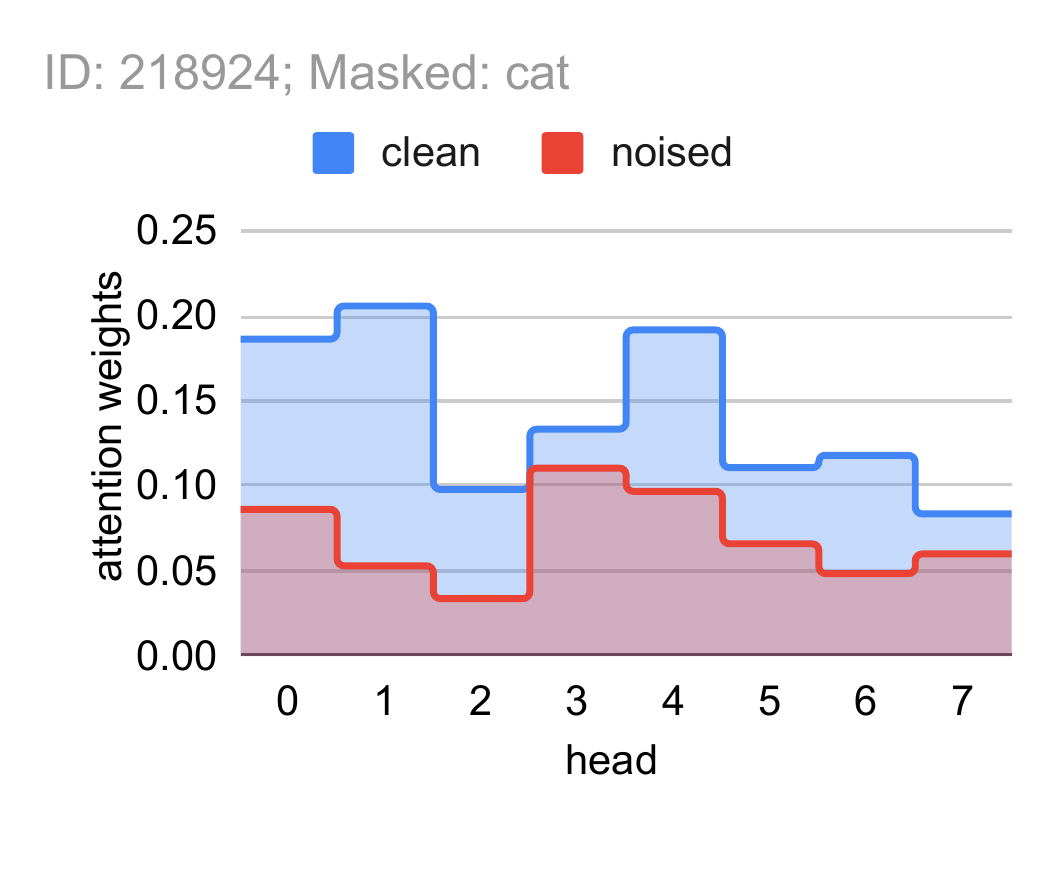}
\end{subfigure}
\vspace{-16pt}
\caption{
\textbf{\textit{(left)}} The attention weights averaged across different attention heads for a noised view drop consistently at each caption generation step.
\textbf{\textit{(center)}} Input image with caption: ``\textit{cat standing in toilet next to a tile floor}''.
\textbf{\textit{(right)}} The attention weights of different attention heads drop consistently at the step of generating the word ``\textit{cat}'', which is masked out in the input image.
}\label{fig:supp:beta5}
\end{figure*}

\begin{figure*}[h!]
\centering
\begin{subfigure}{.39\linewidth}
  \centering
  \includegraphics[width=1.0\linewidth]{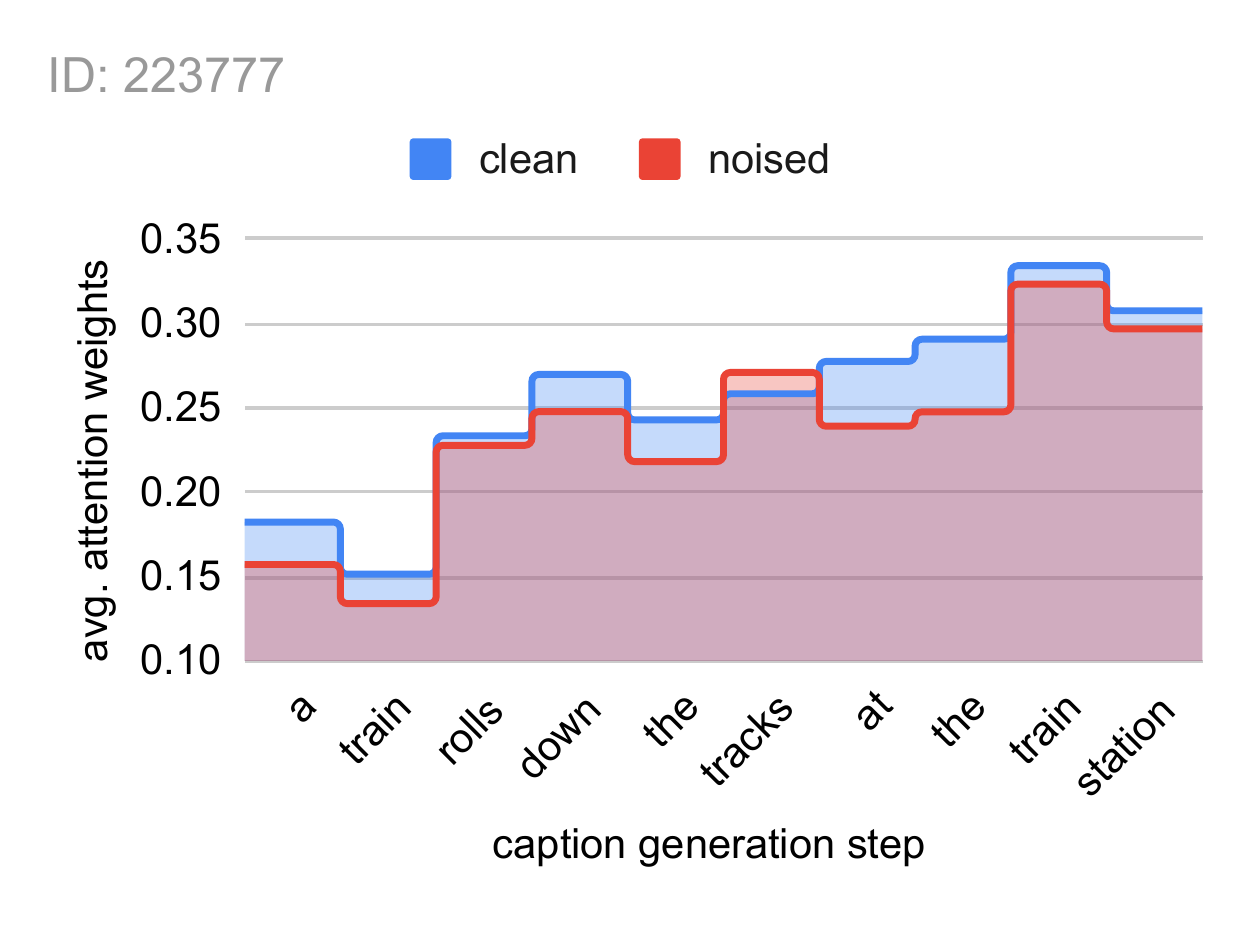}
\end{subfigure}\hfill
\begin{subfigure}{.25\linewidth}
  \centering
  \includegraphics[width=1.0\linewidth]{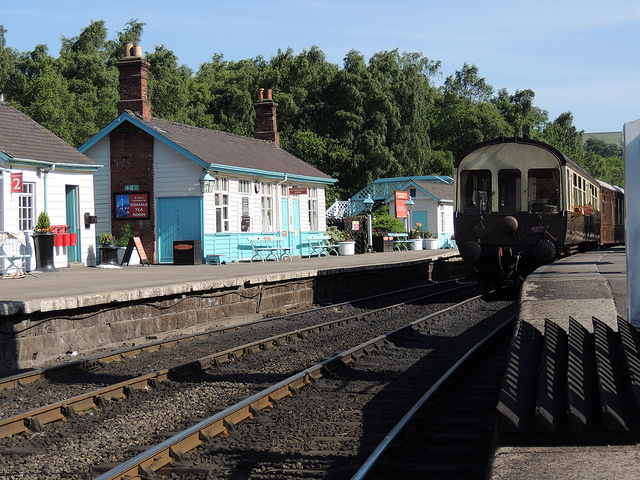}
  \begin{minipage}{.1cm}
    \vfill
  \end{minipage}
\end{subfigure}\hfill
\begin{subfigure}{.34\linewidth}
  \centering
  \includegraphics[width=1.0\linewidth]{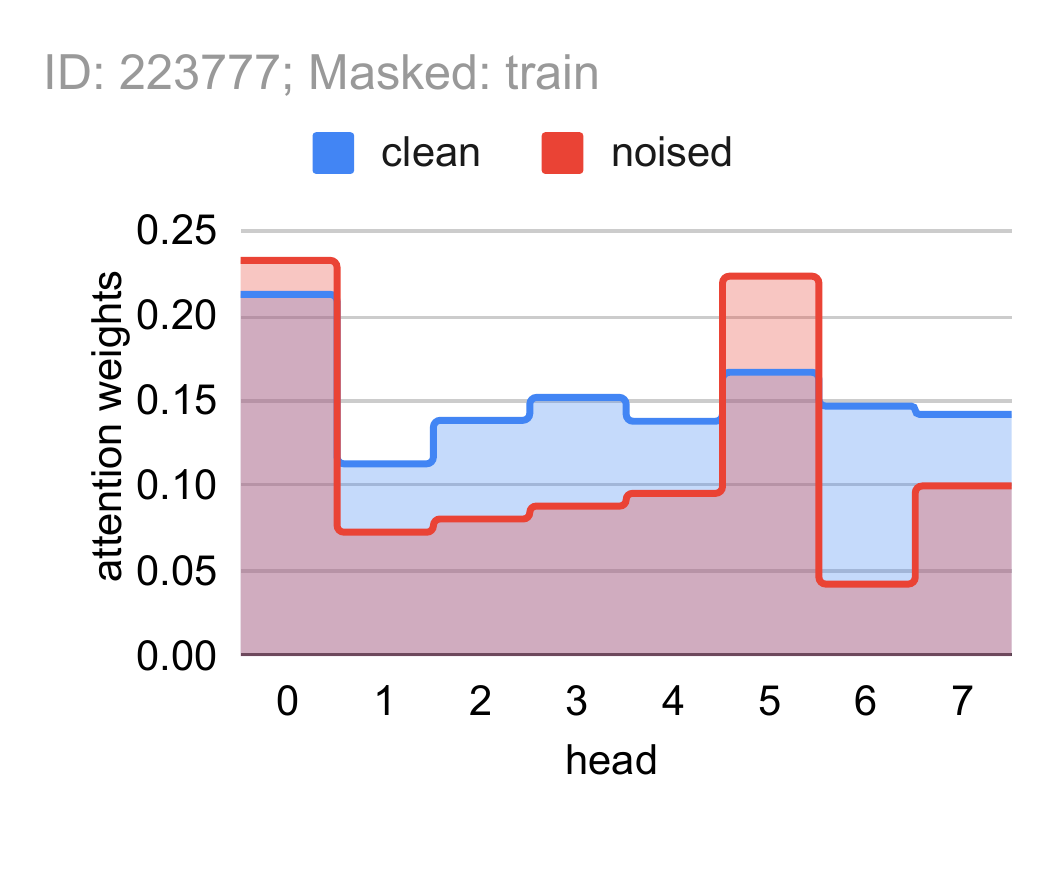}
\end{subfigure}
\vspace{-16pt}
\caption{
\textbf{\textit{(left)}} The attention weights averaged across different attention heads for a noised view drop consistently at each caption generation step.
\textbf{\textit{(center)}} Input image with caption: ``\textit{a train rolls down the tracks at the train station}''.
\textbf{\textit{(right)}} The attention weights of different attention heads drop consistently at the step of generating the word ``\textit{train}'', which is masked out in the input image.
}\label{fig:supp:beta6}
\end{figure*}

We show more examples of how the hierarchical decoder adaptively weighs the encoded views according to their effectiveness for caption generation at the view level and at the word level in Figure~\ref{fig:supp:beta1}-\ref{fig:supp:beta6}.

At the view level (figures on the left), we add noise to a view by randomly zeroing out tokens in a view to make a view \textit{less effective}, and expect a drop of weights toward that noised view.
To measure the weights, we take the multi-head attention weights of $\text{CrossAttn}_\text{Lv2}$ at the last decoder layer and average the attention weights across heads.
Overall, the weights for the noised view drop consistently at each word prediction step compared to the same view \textit{without} added noise.
This means that our hierarchical decoder indeed learns to adaptively weigh the views according to their effectiveness at the view level.

At the word level (figures on the right), we randomly mask out a prominent region of the input image for a view, and expect a drop of the weights toward the masked view at the step of generating the word of that masked region.
To measure the weights, we take the multi-head attention weights of $\text{CrossAttn}_\text{Lv2}$ at the last decoder layer and measure the attention weights of each head at the step of generating the word of that masked region.
Overall, the weights for the masked view drops consistently across all attention heads compared to the same view without masking.
This means that our hierarchical decoder indeed learns to adaptively weigh the input views according to their usefulness at the word level.

~\newpage~\newpage~\newpage~\newpage~\newpage

\end{document}